\RequirePackage{fix-cm}
\documentclass[twocolumn]{svjour3}          
\smartqed  

\usepackage{graphicx}
\usepackage{natbib}
\usepackage{epsfig}
\usepackage{amsmath}
\usepackage{amssymb}
\usepackage{mathtools}
\usepackage{multirow}
\usepackage[pagebackref=true,breaklinks=true,colorlinks=true,citecolor=blue,bookmarks=false]{hyperref}
\usepackage{marvosym}
\usepackage[symbol]{footmisc}
\usepackage[dvipsnames]{xcolor}

\begin{document}
\sloppy 

\title{Estimating 3D Motion and Forces of Human-Object Interactions from Internet Videos}

\author{Zongmian Li\textsuperscript{1,2}         \and
        Jiri Sedlar\textsuperscript{3}         \and
        Justin Carpentier\textsuperscript{1,2}         \and
        Ivan Laptev\textsuperscript{1,2}         \and
        Nicolas Mansard\textsuperscript{4,5}         \and
        Josef Sivic\textsuperscript{3}}


\institute{
\begin{itemize}
\item[\Letter] Zongmian Li
\item[] zongmian.li@inria.fr / zongmian.li@gmail.com
\end{itemize}
\begin{itemize}
\item[\textsuperscript{1}] D\'{e}partement d'informatique de l'ENS, \'{E}cole normale sup\'{e}rieure, CNRS, PSL Research University.
\item[\textsuperscript{2}] Willow project, Inria Paris.
\item[\textsuperscript{3}] Czech Institute of Informatics, Robotics and Cybernetics, Czech Technical University in Prague.
\item[\textsuperscript{4}] LAAS-CNRS, Universit\'{e} de Toulouse, CNRS, Toulouse, France.
\item[\textsuperscript{5}] Artifical and Natural Intelligence Toulouse Insitute (ANITI)
\end{itemize}
}

\date{Received: date / Accepted: date}

\maketitle
 
\begin{abstract}
In this paper, we introduce a method to automatically reconstruct the 3D motion of a person interacting with an object from a single RGB video.
Our method estimates the 3D poses of the person together with the object pose, the contact positions and the contact forces exerted on the human body.
The main contributions of this work are three-fold. 
First, we introduce an approach to jointly estimate the motion and the actuation forces of the person on the manipulated object by modeling contacts and the dynamics of the interactions. 
This is cast as a large-scale trajectory optimization problem.
Second, we develop a method to automatically recognize from the input video the 2D position and timing of contacts between the person and the object or the ground, thereby significantly simplifying the complexity of the optimization. 
Third, we validate our approach on a recent video\,+\,MoCap dataset capturing typical parkour actions, and demonstrate its performance on a new dataset of Internet videos showing people manipulating a variety of tools in unconstrained environments.
\end{abstract}

\section{Introduction}\label{sec:introduction}

\begin{figure}[t]
  \includegraphics[width=\linewidth]{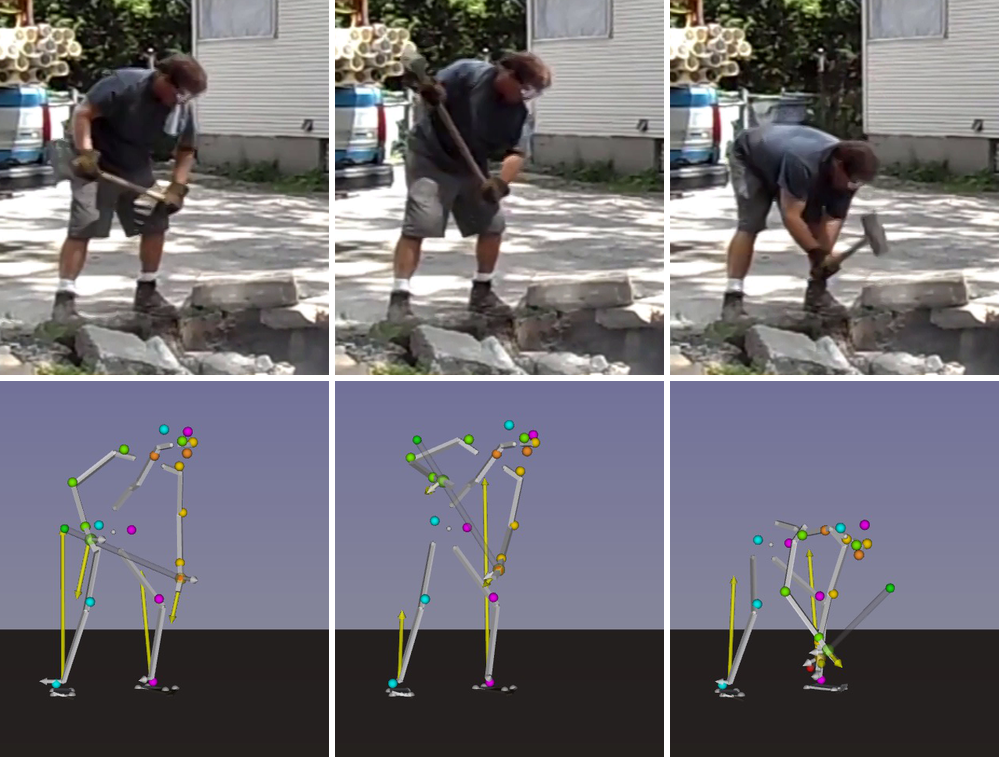}
\caption{
Our method automatically estimates the 3D motion and forces of object manipulation action from a single video.
\textbf{Top row:} sample frames from an input video. \textbf{Bottom row:} 
the estimated person-object 3D motion and 6D contact forces (yellow arrows for linear forces, white arrows for torques).
} 
\label{fig:teaser}
\end{figure}

People can easily learn how to break concrete with a sledgehammer or cut hay using a scythe by observing other people performing such tasks in instructional videos, for example.
They can also easily perform the same task in a different context.
This involves advanced visual intelligence capabilities such as recognizing and interpreting complex person-object interactions that achieve a specific goal. 
Understanding such complex interactions is a key to building autonomous machines that learn how to interact with the physical world by observing people.

This work makes a step in this direction and describes a method to estimate both the 3D motion and the actuation forces of a person manipulating an object given a single unconstrained video as input, as shown in Figure~\ref{fig:teaser}.
This is an extremely challenging task.
First, there are inherent ambiguities in the 2D-to-3D mapping from a single view: multiple 3D human poses correspond to the same 2D input.
Second, human-object interactions often involve contacts, resulting in discontinuities in the motion of the object and the human body part in contact. For example, one must place a hand on the hammer handle before picking the hammer up. The contact motion strongly depends on the physical quantities such as the mass of the object and the contact forces exerted by the hand, which renders modeling of contacts a very difficult task.
Finally, the tools we consider in this work, such as hammer, scythe, or spade, are particularly difficult to recognize due to their thin structure, lack of texture, and frequent occlusions by hands and other parts of human body. 

To address these challenges, we propose a method to jointly estimate the 3D trajectory of both the person and the object by visually recognizing contacts in the video and modeling the dynamics of the interactions.
We focus on rigid stick-like hand tools (e.g.~hammer, barbell, spade, scythe) with no articulation and approximate them as 3D line segments.
Our key idea is that, when a human joint is in contact with an object, the object can be integrated as a constraint on the movement of the human limb.
For example, the hammer in Figure~\ref{fig:teaser} provides a constraint on the relative depth between the person's two hands.
Conversely, 3D positions of the hands in contact with the hammer provide a constraint on the hammer's depth and 3D rotation.
To deal with contact forces, we integrate physics in the estimation by modeling dynamics of the person and the object.
Inspired by recent progress in humanoid locomotion research~\citep{carpentier2018multi}, we formulate person-object trajectory estimation as an optimal control problem given the contact state of each human joint.
We show that contact states can be automatically recognized from the input video using a deep neural network.
Our code and data is available at \url{https://www.di.ens.fr/willow/research/motionforcesfromvideo/}.

\section{Related work}
Here we review the key areas of related work in both computer vision and robotics literature.

\noindent \textbf{Single-view 3D pose estimation } aims to recover the 3D joint configuration of the person from the input image.
Recent human 3D pose estimators either attempt to build a \textit{direct mapping} from image pixels to the 3D joints of the human body or break down the task into \textit{two stages}: estimating pixel coordinates of the joints in the input image and then lifting the 2D skeleton to 3D.
Existing direct approaches either rely on generative models to search the state space for a plausible 3D skeleton that aligns with the image evidence~\citep{sidenbladh2000stochastic,gammeter2008articulated,gall2010optimization} or, more recently, extract deep features from images and learn a regressor from the 2D image to the 3D pose~\citep{hmrKanazawa18,moreno20173d,pavlakos2017coarse,tekin2016direct}.
The models can be further extended to learn 3D human dynamics from 2D in-the-wild video data~\citep{kanazawa2019learning}.

Building on the recent progress in 2D human pose estimation~\citep{newell2016stacked,newell2017associative,insafutdinov2016deepercut,cao2017realtime}, two-stage methods have been shown to be effective~\citep{akhter2015pose,zhou2016sparseness,bogo2016keep,chen20173d} achieving competitive results~\citep{martinez2017simple,xiang2019monocular} on 3D human pose benchmarks~\citep{h36m_pami}. The output can have an impressive level of detail including face deformations and position of individual fingers~\citep{xiang2019monocular}.
To deal with depth ambiguities, these estimators rely on good pose priors, which are either hand-crafted or learnt from large-scale MoCap data~\citep{zhou2016sparseness,bogo2016keep,hmrKanazawa18,kocabas2020vibe}. 
Others have looked at incorporating physical constraints. Examples include incorporating geometric constraints representing the proximity to the ground plane or collisions between different people~\citep{zanfir2018monocular}, or, closer to our work, modelling the dynamics of the human motion and the contacts with the ground~\citep{rempe2020contact, shimada2020physcap}.
However, unlike our work, these methods do not consider explicit physical models for 3D interactions between the person and the handled object.

\noindent \textbf{Understanding human-object interactions} involves both recognition of actions and  modeling of interactions.
In action recognition, most existing approaches that model human-object interactions do not consider 3D, instead model interactions and contacts in the 2D image space~\cite{gupta2009observing,delaitre2011learning,yao2012recognizing,prest2013explicit}.
Recent works in scene understanding~\cite{jiang2013hallucinated,fouhey2014people} consider interactions in 3D but have focused on static scene elements rather than manipulated objects as we do in this work.
Tracking 3D poses of people interacting with the environment has been demonstrated for bipedal walking~\cite{brubaker2007physics,brubaker2009estimating} or in sports scenarios~\cite{videomocap2010}.
However, these works do not consider interactions with objects. Furthermore,~\cite{videomocap2010} requires manual annotation of the input video.

There is also related work on modeling person-object interactions in robotics~\cite{tassa2012} and computer animation~\cite{boulic1990}.
Similarly to people, humanoid robots interact with the environment by creating and breaking contacts~\cite{herdt2010}, for example, during walking.
Typically, generating artificial motion is formulated as an optimal control problem, transcribed into a high-dimensional numerical optimization problem, seeking to minimize an objective function under contact and feasibility constraints~\cite{diehl2006,schultz2010modeling}.
A known difficulty is handling the non-smoothness of the resulting optimization problem introduced by the creation and breaking of contacts~\cite{westervelt2003hybrid}.
Due to this difficulty, the sequence of contacts is often computed separately and not treated as a decision variable in the optimizer~\cite{kuffner2005motion,tonneau2018}.
Recent work has shown that it may be possible to decide both the continuous movement and the contact sequence together, either by implicitly formulating the contact constraints~\cite{posa2014direct} or by using invariances to smooth the resulting optimization problem~\cite{mordatch2012discovery,winkler2018gait}.

In this paper, we take advantage of rigid-body models introduced in robotics and formulate the problem of estimating 3D person-object interactions from monocular video as an optimal control problem under contact constraints.
We overcome the difficulty of contact irregularity by first identifying the contact states from the visual input, and then localizing the contact points in 3D via our trajectory estimator.
This allows us to treat multi-contact sequences (like walking) without manually annotating the contact phases.

\noindent \textbf{Object 3D pose estimation} methods often require depth or RGB-D data as input~\cite{tejani2014latent,doumanoglou2016,hinterstoisser2016going}, which is restrictive since depth information is not always available (e.g.~for outdoor scenes or specular objects), as is the case of our instructional videos. 
Recent work has also attempted to recover object pose from RGB input only~\cite{brachmann2016uncertainty,rad2017bb8,posecnn2017,deepim2018eccv,oberweger2018eccv,grabner2018cvpr,rad2018cvpr}.
However, we found that the performance of these methods is limited for the stick-like objects we consider in this work.
Instead, we recover the 3D pose of the object via localizing and segmenting the object in 2D, and then jointly recovering the 3D trajectory of both the human limbs and the object. As a result, both the object and the human pose help each other to improve their joint 3D trajectory by leveraging the contact constraints.

\noindent \textbf{Instructional videos.}
Our work is also related to recent efforts in learning from Internet instructional videos~\cite{malmaud2015s,Alayrac16unsupervised} that aim to segment input videos into clips containing consistent actions. 
In contrast, we focus on extracting a detailed representation of the object manipulation in the form of a 3D person-object trajectory with contacts and underlying interaction forces.

\begin{figure*}[t]
    \centering
    \includegraphics[width=\textwidth]{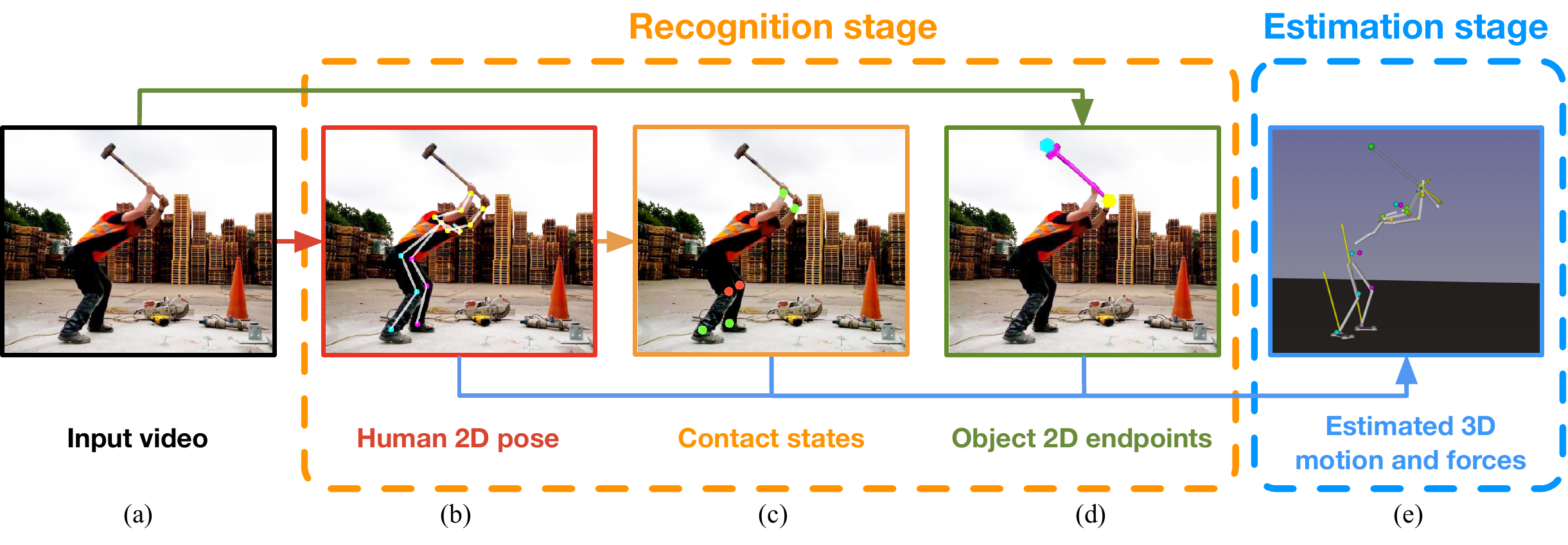}
    \caption{Overview of the proposed method. In recognition stage (orange box, b-d), the system estimates from the input video (a) the locations of person's 2D joints (b), the locations of 2D endpoints of the tool (d), and contact states of the individual joints (c).
    The human joints and the object endpoints are visualized as colored dots in the image.
    Human joints recognized as in contact are shown in green in (c), joints not in contact in red.
    In estimation stage (blue box, (e)), these image measurements are fused in a trajectory estimator to recover the human and object 3D motion together with the contact positions and forces (shown as yellow arrows).}
    \label{fig:system_model_scheme_1}
\end{figure*}

\section{Approach overview}
We are given a video clip of a person manipulating an object or in another way interacting with the scene. 
Our approach, illustrated in Figure~\ref{fig:system_model_scheme_1}, receives as input a sequence of frames and automatically outputs the 3D trajectories of the human body, the manipulated object, and the ground plane. At the same time, it localizes the contact points and recovers the contact forces that actuate the motion of the person and the object.
Our approach proceeds along two stages. 
In the first stage, \textit{the recognition stage}, we extract 2D measurements from the input video. These consist of 2D locations of human joints, 2D locations of a small number of predefined object endpoints, and contact states of selected joints over the course of the video. 
In the second stage, \textit{the estimation stage}, these image measurements are then fused in order to estimate the 3D motion, 3D contacts, and the controlling forces of both the person and the object.
The person and object trajectories, contact positions, and contact forces are jointly constrained by our carefully designed contact motion model, force model, and dynamics equations.

The proposed problem is difficult, yet feasible to solve under a number of reasonable assumptions on the physical properties of the person, the manipulated object and the scene.
First of all, we assume that there is at most one person that appears in the input video.
We adopt the mass properties of the full-body anatomical human model described in~\citep{maldonado-models}.
This model captures the body weight statistics of an average human adult.
Our approach applies the same body mass distribution to any input video. 
If there is an object manipulated by the person, we assume that the object is rigid, non-articulated and has a stick-like shape.
We apply a single object mass distribution to any input video with the same type of object.
For example, we assume that all sledgehammers share the same head weight. 
Our method can also handle input videos without the manipulated object.
In this case, we only model contacts between the person and the ground. 
We further assume the camera is static with canonical (or known) intrinsic parameters.
Most body joints, especially the ones that may interact with the environment (e.g. hands, feet, knees, etc) should be visible at least in a short period of time in the input video.
We assume that the gravity is perpendicular to the ground plane, but the model can be tuned to fit other cases such as a sloping ground.
In the subsequent sections, we will include an object model in our formulation, but as discussed above the object is not necessary for the model to be applied.

In the following, we start in Section~\ref{sec:main_stage} by describing the estimation stage giving details of the formulation as an optimal control problem.
Then, in Section~\ref{sec:extract_2d_measurements} we give details of the recognition stage including 2D human pose estimation, contact recognition, and object 2D endpoint estimation.
Finally, we describe results including the failure modes in Section~\ref{sec:results}.

\section{Estimating person-object trajectory under contact and dynamics constraints}\label{sec:main_stage}
We assume that we are provided with a video clip of duration $T$ depicting a human subject manipulating an object.
We encode the 3D poses of the human and the object, including joint translations and rotations, in the configuration vectors $q^\mathrm{h}$ and $q^\mathrm{o}$, for the human and the object respectively.
We define a constant set of $K$ contact points between the human body and the object (or the ground plane).
Each contact point corresponds to a human segment, and is activated whenever that human segment is recognized as in contact.
At each contact point, we define a contact force $f_k$, whose value is non-zero whenever the contact point $k$ is active.
The state of the complete dynamical system is then obtained by concatenating the human and the object joint configurations $q$ and velocities $\dot{q}$ as 
$x \coloneqq \left(q^\mathrm{h}, q^\mathrm{o}, \dot{q}^\mathrm{h}, \dot{q}^\mathrm{o}\right)$.
Let $\tau^\mathrm{h}_\mathrm{m}$ be the joint torque vector describing the actuation by human muscles. This is a $n_q-6$ dimensional vector where $n_q$ is the dimension of the human body configuration vector. We define the control variable $u$ as the combination of the joint torque vector together with the contact forces at the $K$ contact point, $u \coloneqq \left(\tau^\mathrm{h}_\mathrm{m}, f_k, k=1,...,K\right)$.
To deal with sliding contacts, we further define a contact state $c$ that consists of the relative positions of all the contact points with respect to the object (or ground) in the 3D space.

Our goal is two-fold. We wish to (i) estimate smooth and consistent human-object and contact trajectories $\underline{x}$ and $\underline{c}$, while (ii) recovering the control $\underline{u}$ which gives rise to the observed motion\footnote[2]{In this paper, trajectories are denoted as underlined variables, e.g. $\underline{x},\underline{u}~\text{or}~\underline{c}$.}.
This is achieved by jointly optimizing the 3D trajectory $\underline{x}$, contacts $\underline{c}$, and control $\underline{u}$ given the measurements (2D positions of human joints and object endpoints together with contact states of human joints) obtained from the input video. 
The intuition is that the human and the object's 3D poses should match their respective projections in the image while their 3D motion is linked together by the recognized contact points and the corresponding contact forces.
In detail, we formulate person-object interaction estimation as an optimal estimation problem with contact and dynamics constraints:
\begin{align}
    \underset{\underline{x},\underline{u},\underline{c}}{\text{minimize}} &\quad \sum_{e \in \{\mathrm{h}, \mathrm{o}\}}{\int_{0}^{T}{l^{e}\left(x, u, c\right)\mathrm{d}t}}, \label{eq:general_problem}\\
    \text{subject to} &\quad  \kappa(x, c) = 0 \quad \text{(contact motion model)}, \label{eq:contact_motion_model}\\
    & \quad \dot{x} = f\left(x, c, u\right)  \quad \text{(full-body dynamics)}, \label{eq:full_body_dynamics}\\
    &\quad u \in \mathcal{U}\quad \text{(force model)},\label{eq:force_model}
\end{align}
where $e$ denotes either `$\mathrm{h}$' (human) or `$\mathrm{o}$' (object), and the constraints \eqref{eq:contact_motion_model}-\eqref{eq:force_model} must hold for all $t\in[0,T]$.
The loss function $l^e$ is a weighted sum of multiple costs capturing (i) the data term measuring simultaneously the consistency between the observed and re-projected 2D joint and object endpoint positions and the discrepancy of the estimated 3D joint positions with respect to some reference positions, (ii) the prior on the human 3D poses, (iii) the physical plausibility of the motion and (iv) the temporal smoothness of the estimated trajectory.
Next, we describe these cost terms as well as the insights leading to their design choices.
For simplicity, we ignore the superscript $e$ when introducing a cost term that exists for both the human $l^\mathrm{h}$ and the object $l^\mathrm{o}$ component of the loss.
We describe the individual terms using continuous time notation as used in the overall problem formulation~\eqref{eq:general_problem}. 
A discrete version of the problem as well as the optimization and implementation details are relegated to Section~\ref{sec:optimization}.

\subsection{Data term: enforcing 2D and 3D consistency}
Given the 2D locations of human joints and object endpoints predicted from image, we wish to optimize a 3D pose trajectory that consolidates these 2D measurements.
This is done by minimizing the re-projection error of the estimated 3D human joints and 3D object endpoints with respect to the 2D measurements obtained in each video frame.
In detail, let $j=1,...,N$ be human joints or object endpoints and $p^{\mathrm{2D}}_j$ their 2D position observed in the image. 
We minimize the 2D consistency loss $l_\mathrm{2D}$:
\begin{equation}
    l_\mathrm{2D} =  \sum_j\rho\left(p^{\mathrm{2D}}_{j} - P_\mathrm{cam}(p_{j}(q))\right) \label{eq:data-2d},
\end{equation}
where $P_\mathrm{cam}$ is the camera projection matrix and $p_j$ the 3D position of joint or object endpoint $j$ induced by the  person-object configuration vector $q$.
To deal with outliers, we use the robust Huber loss, denoted by $\rho$.

In addition, we employ a direct 3D consistency loss if a reference 3D pose trajectory is available:
\begin{align}
    l_\mathrm{3D} =  \sum_j\rho\left(p^{\mathrm{3D}}_{j} - p_{j}(q)\right) \label{eq:data-3d},
\end{align}
where $p^{\mathrm{3D}}_{j}$ denotes the reference 3D position of joint $j$.
In our case, the reference human 3D poses are computed using the HMR estimator \citep{hmrKanazawa18}.
But it is possible to use other pose estimators instead.

In practice, we find that minimizing a weighted sum of the 2D and 3D consistency losses achieves good performance. 
The data term is finally expressed as:
\begin{align}
    l_\mathrm{data} = w_\mathrm{2D}l_\mathrm{2D} + w_\mathrm{3D}l_\mathrm{3D} \label{eq:data},
\end{align}
where $w_\mathrm{2D}$ and $w_\mathrm{2D}$ are non-negative scalars.

\subsection{Prior on 3D human poses}
\label{sec:pose_prior}
A single 2D skeleton can be a projection of multiple 3D poses, many of which are unnatural or impossible exceeding the human joint limits.
To resolve this, we incorporate into the human loss function $l^\mathrm{h}$ a pose prior similar to~\citet{bogo2016keep}.
The pose prior is obtained by fitting the SMPL human model \citep{loper2015smpl} to the CMU MoCap dataset using MoSh \citep{Loper:SIGASIA:2014} and fitting a Gaussian Mixture Model (GMM) to the resulting SMPL 3D poses.
We map our human configuration vector $q^\mathrm{h}$ to a SMPL pose vector $\theta$ and compute the likelihood under the pre-trained GMM
\begin{align}
l^\mathrm{h}_\mathrm{pose} &= -\log\left(p(q^\mathrm{h}; \text{GMM})\right).\label{eq:human_pose}
\end{align}
During optimization, $l^\mathrm{h}_\mathrm{pose}$ is minimized in order to favor more plausible human poses against rare or impossible ones.

\subsection{Physical plausibility of the motion}\label{sec:physical_plausibility}
Human-object interactions involve contacts coupled with interaction forces, which are not included in the data-driven cost terms \eqref{eq:data} and \eqref{eq:human_pose}.
Modeling contacts and physics is thus important to reconstruct object manipulation actions from the input video.
Next, we outline models for describing the motion of the contacts and the forces at the contact points. Finally, the contact motions and forces, together with the system state $\underline{x}$, are linked by the laws of mechanics via the dynamics equations, which constrain the estimated person-object interaction. This full body dynamics constraint is detailed at the end of this subsection.

\paragraph{Contact motions.}
In the recognition stage, our contact recognizer predicts, given a human joint (for example, left hand, denoted by $j$), a sequence of contact states $\delta_j: t \longrightarrow \{1,0\}$.
Similarly to \cite{carpentier2018multi}, we call a \textit{contact phase} any time segment in which $j$ is in contact, i.e., $\delta_j=1$.
Our key idea is that the 3D distance between human joint $j$ and the active contact point on the object (denoted by $k$) should remain zero during a contact phase:
\begin{align}
     \left\|p^\mathrm{h}_j(q^\mathrm{h}) - p^\mathrm{c}_k(x, c)\right\|=0\quad  \text{(point contact)},\label{eq:point_contact}
\end{align}
where $p^\mathrm{h}_{j}$ and $p^\mathrm{c}_k$ are the 3D positions of joint $j$ and object contact point $k$, respectively. Note that position of the object contact point $p^\mathrm{c}_k(x, c)$ depends on the state vector $x$ describing the human-object configuration and the relative position $c$ of the contact along the object. 
The position of contact $p^\mathrm{c}_k$ is subject to a feasible range denoted by $\mathcal{C}$.
For stick-like objects such as hammer, $\mathcal{C}$ is approximately the 3D line segment representing the handle. For the ground, the feasible range $\mathcal{C}$ is a 3D plane. 
In practice, we implement $p^\mathrm{c}_k\in \mathcal{C}$ by putting a constraint on the trajectory of relative contact positions $\underline{c}$.

Equation~\eqref{eq:point_contact} applies to most common cases where the contact area can be modeled as a point.
Examples include the hand-handle contact and the knee-ground contact.
To model the \textit{planar contact} between the human sole and ground, we approximate each sole surface as a planar polygon with four vertices, and apply the point contact model at each vertex.
In our human model, each sole is attached to its parent ankle joint, and therefore the four vertex contact points of the sole are active when $\delta_\mathrm{ankle}=1$.

The resulting overall contact motion function $\kappa$ in problem \eqref{eq:general_problem} is obtained by unifying the point and the planar contact models:
\begin{align}
\kappa(x, c)=\sum_{j}\sum_{k\in \phi(j)}\delta_j\left\|T^{(kj)} \left(p^\mathrm{h}_j(q^\mathrm{h})\right) - p^\mathrm{c}_k(x, c)\right\|, \label{eq:contact_model}
\end{align}
where the external sum is over all human joints.
The internal sum is over the set of active object contact points mapped to their corresponding human joint $j$ by mapping $\phi(j)$.
The mapping $T^{(kj)}$ translates the position of an ankle joint $j$ to its corresponding $k$-th sole vertex; it is an identity mapping for non-ankle joints.

\paragraph{Contact forces.}
During a contact phase of the human joint $j$, the environment exerts a contact force $f_k$ on each of the active contact points in $\phi(j)$.
$f_k$ is always expressed in contact point $k$'s local coordinate frame.
We distinguish two types of contact forces: (i) 6D spatial forces exerted by objects and (ii) 3D linear forces due to ground friction.
In the case of object contact, $f_k$ is an unconstrained 6D spatial force with 3D linear force and 3D moment.
In the case of ground friction, $f_k$ is constrained to lie inside a 3D friction cone $\mathcal{K}^3$ (also known as the quadratic Lorentz ``ice-cream'' cone \citep{carpentier2018multi}) characterized by a positive friction coefficient $\mu$.
In practice, we approximate $\mathcal{K}^3$ by a 3D pyramid spanned by a basis of $N=4$ generators, which allows us to represent $f_k$ as the convex combination $f_k = \sum_{n=1}^{N}{\lambda_{kn}g^{(3)}_n}$, where $\lambda_{kn}\geq 0$ and $g^{(3)}_n$ with $n=1,2,3,4$ are the 3D generators of the contact force.
We sum the contact forces induced by the four sole-ground contact points and express a unified contact force in the ankle's frame:
\begin{align}
    f_j=\sum_{k=1}^4
    \begin{pmatrix}
        f_k \\ 
        p_k\times f_k
    \end{pmatrix}
    =\sum_{k=1}^{4}\sum_{n=1}^{N}\lambda_{jkn}g^{(6)}_{kn},
\end{align}
where $p_k$ is the position of contact point $k$ expressed in joint $j$'s (left/right ankle) frame, $\times$ is the cross product operator, $\lambda_{jkn}\geq 0$, and $g^{(6)}_{kn}$ are the 6D generators of $f_j$.
Please see Appendix \ref{appendix:ground_force_generators} for additional details including the expressions of $g^{(3)}_{n}$ and $g^{(6)}_{kn}$.

\paragraph{Full body dynamics.}
The full-body movement of the person and the manipulated object is described by the Lagrange dynamics equation:
\begin{align}
    M(q)\ddot{q} + b(q, \dot{q}) = g(q) + \tau, \label{eq:dynamics_equation}
\end{align}
where $M$ is the generalized mass matrix, $b$ covers the centrifugal and Coriolis effects, $g$ is the generalized gravity vector and $\tau$ represents the joint torque contributions.
$\dot{q}$ and $\ddot{q}$ are the joint velocities and joint accelerations, respectively.
Note that \eqref{eq:dynamics_equation} is a unified equation which applies to both human and object dynamics, hence we drop the superscript $e$ here. Only the expression of the joint torque $\tau$ differs between the human and the object and we give the two expressions next. 

For human, it is the sum of two contributions: the first one corresponds to the internal joint torques (exerted by the muscles for instance) and the second one comes from the contact forces:
\begin{align}
    \tau^\mathrm{h} =
    \begin{pmatrix}
        \mathbf{0}_6 \\ 
        \tau^\mathrm{h}_\mathrm{m}
    \end{pmatrix}
    + \sum_{k=1}^K \left(J^\mathrm{h}_k\right)^Tf_k, \label{eq:human_torque}
\end{align}
where $\tau^\mathrm{h}_\mathrm{m}$ is the human joint torque exerted by muscles, $f_k$ is the contact force at contact point $k$ and $J^\mathrm{h}_k$ is the Jacobian mapping human joint velocities $\dot{q}^\mathrm{h}$ to the Cartesian velocity of contact point $k$ expressed in $k$'s local frame.
Let $n^\mathrm{h}_q$ denote the dimension of $q^\mathrm{h}$, $\dot{q}^\mathrm{h}$ and $\ddot{q}^\mathrm{h}$, then $\tau^\mathrm{h}_\mathrm{m}$ and $J^\mathrm{h}_k$ are of dimension $n_q^h-6$ and $3\times n_q^h$, respectively.
We model the human body and the object as two free-floating base systems.
In the case of human body, the six first entries in the configuration vector $q$ correspond to the 6D pose of the free-floating base (translation + orientation), which is not actuated by any internal actuators such as human muscles.
This constraint is taken into consideration by adding the zeros in Eq.\,\eqref{eq:human_torque}.

In the case of the manipulated object, there is no actuation other than the contact forces exerted by the human.
Therefore, the object torque is expressed as
\begin{align}
    \tau^\mathrm{o} =
    -\sum_{\text{object contact }k} \left(J^\mathrm{o}_k\right)^Tf_k, \label{eq:object_torque}
\end{align}
where the sum is over the object contact points, $f_k$ is the contact force, and $J^\mathrm{o}_k$ denotes the object Jacobian, which maps from the object joint velocities $\dot{q}^\mathrm{o}$ to the Cartesian velocity of the object contact point $k$ expressed in $k$'s local frame.
$J^\mathrm{o}_k$ is a $3\times n^\mathrm{o}_q$ matrix where $n^\mathrm{o}_q$ is the dimension of object configuration vectors $q^\mathrm{o}$, $\dot{q}^\mathrm{o}$ and $\ddot{q}^\mathrm{o}$.

We concatenate the dynamics equations of both human and object to form the overall dynamics in Eq.\,\eqref{eq:full_body_dynamics} in problem \eqref{eq:general_problem}, and include a \textit{muscle torque} term  
$l^\mathrm{h}_\mathrm{torque} = \|\tau^\mathrm{h}_\mathrm{m}\|^2$ in the overall cost. Minimizing the muscle torque acts as a regularization over the energy consumption of the human body.

\subsection{Enforcing the trajectory smoothness}
\label{sec:smoothness}

\paragraph{Regularizing human and object motion.}
Taking advantage of the temporal continuity of video, we minimize the sum of squared 3D joint velocities and accelerations to improve the smoothness of the person and object motion and to remove incorrect 2D poses.
We include the following \textit{motion smoothing} term to the human and object loss in \eqref{eq:general_problem}:
\begin{align}
  l_\mathrm{smooth} = \sum_{j}{\left(\left\|\nu_j(q, \dot{q})\right\|^2 + \left\|\alpha_j(q, \dot{q}, \ddot{q})\right\|^2\right)},
  \label{eq:cost_smooth}
\end{align}
where $\nu_j$ and $\alpha_j$ are the spatial velocity and the spatial acceleration\footnote[3]{Spatial velocities (accelerations) are minimal and unified representations of linear and angular velocities (accelerations) of a rigid body \citep{featherstone2014rigid}. They are of dimension 6.} of joint $j$, respectively.
In the case of object, $j$ represents an endpoint on the object.
By minimizing $l_\mathrm{smooth}$, both the linear and angular movements of each joint/endpoint are smoothed simultaneously.

\paragraph{Regularizing contact motion and forces.}

In addition to regularizing the motion of the joints, we also regularize the contact states and control by minimizing the velocity of the contact points, the temporal variation of the contact forces and the magnitude of the contact forces.
The is implemented by including the following \textit{contact smoothing} term in the cost function in problem \eqref{eq:general_problem}:
\begin{align}
    l^\mathrm{c}_\mathrm{smooth}
    = \sum_{j}\sum_{k\in\phi(j)}\left(\omega_k\|\dot{c}_k\|^2+\gamma_k\|\dot{f}_k\|^2 + \zeta_k\|f_k\|^2\right), \label{eq:contact_smooth}
\end{align}
where $\dot{c}_k$ and $\dot{f}_k$ represent, respectively, the temporal variation of the position and the contact force at contact point $k$. $f_k$ is the contact force at contact point $k$. $\omega_k$, $\gamma_k$ and $\zeta_k$ are scalar weights of the regularization terms. 
Note that some contact points, for example the four contact points of the human sole during the sole-ground contact, should remain fixed with respect to the object or the ground during the contact phase.
To tackle this, we use a higher $\omega_k$ for sole contact points to prevent the foot sole form sliding.
We also found important to use higher $\zeta_k$ for hand contact forces and smaller $\zeta_k$ for ground contact forces to favor larger ground contact forces when both hand and ground contacts are recognized.

\subsection{Optimization}\label{sec:optimization}

\paragraph{Conversion to a numerical optimization problem.}
We convert the continuous problem~\eqref{eq:general_problem} into a discrete nonlinear optimization problem using the collocation approach~\citep{biegler2010nonlinear_chap10}.
All trajectories are discretized and the constraints~\eqref{eq:contact_motion_model}, \eqref{eq:full_body_dynamics}, \eqref{eq:force_model} are only enforced on the ``collocation'' nodes of a time grid matching the discrete sequence of video frames.
The optimization variables are the sequence of human and object poses $[ x_0 ... x_T ]$, torque and force controls $[ u_1 ... u_T ]$, contact locations $[c_0 ... c_T]$, and the ground plane.
We replace the integral in the objective function by a sum over video frames, and rewrite the cost and constraint terms which include derivatives of the state (e.g. joint accelerations) by approximating the derivatives with the backward finite difference scheme (e.g. $a_t := (v_{t} - v_{t-1}) / \Delta t $, with $\Delta t$ the duration between two video frames).
The resulting problem is nonlinear, constrained and sparse (due to the sequential structure of trajectory optimization).

\paragraph{Problem sparsity.}
The problem after discretization becomes a large, sparse and non-linear optimization problem.
This is because the discretized objective function becomes a sum of terms that each depend on one time sample (denoted by $i$) and a subset of the variables $[x_i,u_i,c_i]$ corresponding to $i$.
Only a few regularization terms, e.g. the motion smoothing term \eqref{eq:cost_smooth} and the contact smoothing term \eqref{eq:contact_smooth}, may depend on two or three successive frames.
The problem sparsity is important to take into account, as it significantly reduces the complexity of computation from $\mathcal{O}(T^3)$ (without sparsity) to $\mathcal{O}(T)$ (using the problem sparsity).

\paragraph{Solving the problem.}
We solve the problem using the Levenberg-Marquardt algorithm.
We rely on the Ceres solver~\citep{ceres-solver}, which is dedicated to solving sparse estimation problems (e.g.~bundle adjustment~\citep{triggs1999bundle}), and on the Pinocchio software~\citep{carpentier2019pinocchio,pinocchioweb} for the efficient computation of kinematic and dynamic quantities and their derivatives~\citep{carpentier2018analytical}.
As Ceres solver only allows to define bound constraints, hence we implement our nonlinear constraints as penalties in the cost function.

\paragraph{Multi-stage optimization.} 
In practice, we find that solving the optimization problem all at once usually leads to poor local minima. Instead we design a multi-stage optimization strategy taking inspiration in multi-stage optimization used for planning motion of humanoid robots~\cite{tonneau2018efficient,carpentier2017multi}. 
In detail, we solve a cascade of sub-problems composed of four stages.

In stage 1, we solve the discretized version of problem~\eqref{eq:general_problem} only for the person's kinematic variables ($q^\mathrm{h}$, $\dot{q}^\mathrm{h}$, $\ddot{q}^\mathrm{h}$) by ``freezing'' all variables and constraints related to the object, the ground plane, and the dynamics in Equations~\eqref{eq:full_body_dynamics} and~\eqref{eq:force_model}.
This gives us a rough estimate of the person's 3D trajectory.

In stage 2, we recover the 3D position of the ground plane given the estimated 3D trajectory of the person and the contact states recognized from the video sequence.
In detail, we ``unfreeze'' the 3D position $q^\mathrm{g}$ ground plane and jointly solve for the trajectory of the person $q^\mathrm{h}$ and the position of the ground plane $q^\mathrm{g}$.

Stage 3 is dedicated to initializing the object's 3D trajectory.
This is achieved by solving for the object's kinematic variables ($q^\mathrm{o}$, $\dot{q}^\mathrm{o}$, $\ddot{q}^\mathrm{o}$) under the contact constraints, while keeping the other variables fixed.
Note that the location of the manipulated object varies significantly across the Handtool dataset.
To address this, we sample four initialization options with different pre-defined 3D object orientations.
We run stage 3 of the optimization for each initialization and pick among the four resulting solutions the one with the lowest cost.

Finally, in stage 4, we solve for the complete set of kinematic and control variables all at once, starting from the values provided by the previous stages.
It is possible to continue improving the solution by pursing the aforementioned  alternative descent scheme, but we found that a single pass was already sufficient to obtain good results.

\paragraph{Setting hyper-parameters.}
Hyper-parameters of our trajectory estimator, including the weights used for the cost terms, the camera model, the number of iterations, etc., are determined by following a combination of manual adjustment and a grid search: given a parameter of interest and a search grid, we run the optimization on a set of validation videos with known ground-truth 3D motion, evaluate the joint errors at every grid point, and update the hyper-parameter with the value leading to the lowest error.
The same process is repeated in an iterative manner for the different hyper-parameters until the model outputs reasonable results on all the validation videos.

\paragraph{Run time.} 
    We report run time of trajectory optimization on a MacBook Pro 2016 (with 2.9GHz Intel Core i5 and 8GB memory).
    The optimization takes on average 3.23 seconds per frame. In detail, the four stages of the optimization, from stage 1 to stage 4, take on average 0.40, 0.02, 0.31 and 2.50 seconds per frame, respectively. When the pose of the object is not modeled, which is the case of one of our datasets introduced in the experimental section, the optimization is faster as stage 3 is skipped. 
    By default, the optimization is run on the whole input video (around 100 frames in our datasets). 
    We also provide an interface for running the optimization in a sliding window manner, which allows applying our method on longer videos.

\section{Extracting 2D measurements from video}\label{sec:extract_2d_measurements}
In this section, we describe how 2D measurements are extracted from the input video frames during the first, recognition stage of our system.
In particular, we extract the 2D human joint positions, the 2D object endpoint positions and the contact states of human joints.

\paragraph{Estimating 2D positions of human joints.}\label{sec:human_2d_pose}
We use the state-of-the-art Openpose~\citep{cao2017realtime} human 2D pose estimator, which achieved excellent performance on the MPII Multi-Person benchmark~\citep{andriluka20142d}.
Taking a pre-trained Openpose model, we do a forward pass on the input video in a frame-by-frame manner to obtain an estimate of the 2D trajectory of human joints, $p^\mathrm{h,2D}_j$.

\paragraph{Recognizing contacts.}\label{sec:contact_recognition}
\begin{figure}[t]
    \centering
    \includegraphics[width=0.49\textwidth]{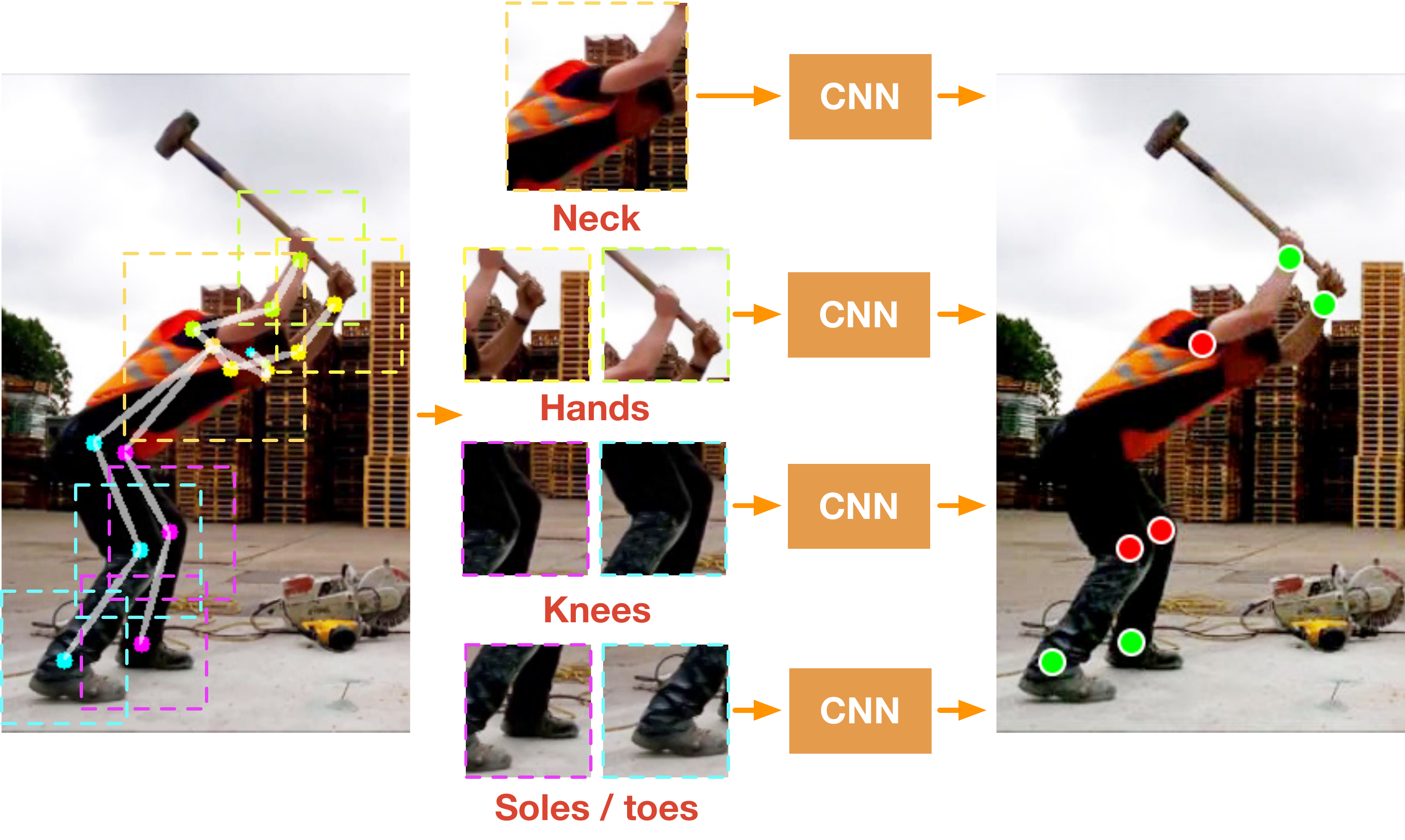}
    \caption{The main contact recognition steps. 
    Given estimated 2D human joints, we crop image patches around a set of joints of interest, which includes neck, hands, knees, foot soles and toes.
    Based on the type of human joint, we feed each image patch to the corresponding CNN to predict whether the joint appearing in the patch is in contact (shown in green on the right) or not (shown in red) with the environment.
    }
    \label{fig:contact_recognizer}
\end{figure}

We wish to recognize and localize contact points between the person and the manipulated object or the ground. 
This is a challenging task due to the large appearance variation of the contact events in the video. 
However, we demonstrate here that a good performance can be achieved by training a contact recognition CNN module from manually annotated contact data that combine both still images and videos harvested from the Internet. 
In detail, the contact recognizer operates on the 2D human joints predicted by Openpose.
As shown in Figure~\ref{fig:contact_recognizer}, given 2D joints at video frame $i$, we crop fixed-size image patches around a set of joints of interest, which may be in contact with an object or ground.
Based on the type of human joint, we feed each image patch to the corresponding CNN to predict whether the joint appearing in the patch is in contact or not.
The output of the contact recognizer is a sequence $\delta_{ji}$ encoding the contact states of human joint $j$ at video frame $i$, i.e.\  $\delta_{ji}=1$ if joint $j$ is in contact at frame $i$ and zero otherwise.
Note that $\delta_{ji}$ is the discretized version of the contact state trajectory $\delta_j$  presented in Sec.~\ref{sec:main_stage}.

Our contact recognition CNNs are built by replacing the last layer of an ImageNet pre-trained Resnet model \citep{he2016deep} with a fully connected layer that has a binary output.
We have trained separate models for five types of joints: hands, knees, foot soles, toes, and neck. 
To construct the training data, we collect still images of people manipulating tools using Google image search.
We also collect short video clips of people manipulating tools from Youtube in order to also have non-contact examples.
We run Openpose pose estimator on this data, crop patches around the 2D joints, and annotate the resulting dataset with contact states.

\paragraph{Estimating 2D object pose.}\label{sec:object_2d_endpoints}
The objective is to estimate the 2D position of the manipulated object in each video frame.
To achieve this, we build on instance segmentation, computed by Mask R-CNN~\citep{MaskRCNN}.
We train Mask R-CNN separately for each object class (i.e., barbell, hammer, scythe and spade) and apply it to the corresponding Handtool dataset videos.
Using the inferred segmentation masks and bounding boxes, we estimate the 2D location of the object endpoints (i.e.\ its two extremities) in each frame.
The resulting 2D endpoint coordinates are used as an input to the trajectory optimizer.
Details are given next.

In order to generate training data for the instance segmentation, we used two different approaches.
In the case of barbell, hammer and scythe, we created a 3D model for each object class (i.e. one model for all barbell instances, for example), roughly approximating the shape of the corresponding object instances in the Handtool dataset videos, and computed the mask of the model shape in 2D from multiple viewpoints using a perspective camera.
For spade, we collected a small number (13) of still images capturing different instances of person-spade manipulation similar to those in the Handtool dataset, and annotated 2D masks of the spade in them.
Then we augmented the resulting 2D shape masks to train a separate Mask R-CNN model for each object class.
In order to handle the variation of object poses in the videos, we augmented the training set by random 2D geometric transformations (translation, rotation, scale, flip).
In addition, to handle the intra-class variation of instance surface appearance as well as changes caused by illumination,
we applied domain randomization \citep{loing2018,tobin2017corr}: the geometrically transformed 2D mask was filled with a random (foreground) image and pasted on another random (background) image; the random images were taken from the MS COCO dataset \citep{COCO}.
Starting with a Mask R-CNN \citep{matterport_maskrcnn_2017} model pre-trained on the MS COCO dataset, we train a separate model for each object class by fine-tuning the head layers using the corresponding augmented training set.

\begin{figure}[t]
    \centering
    \includegraphics[width=0.47\textwidth]{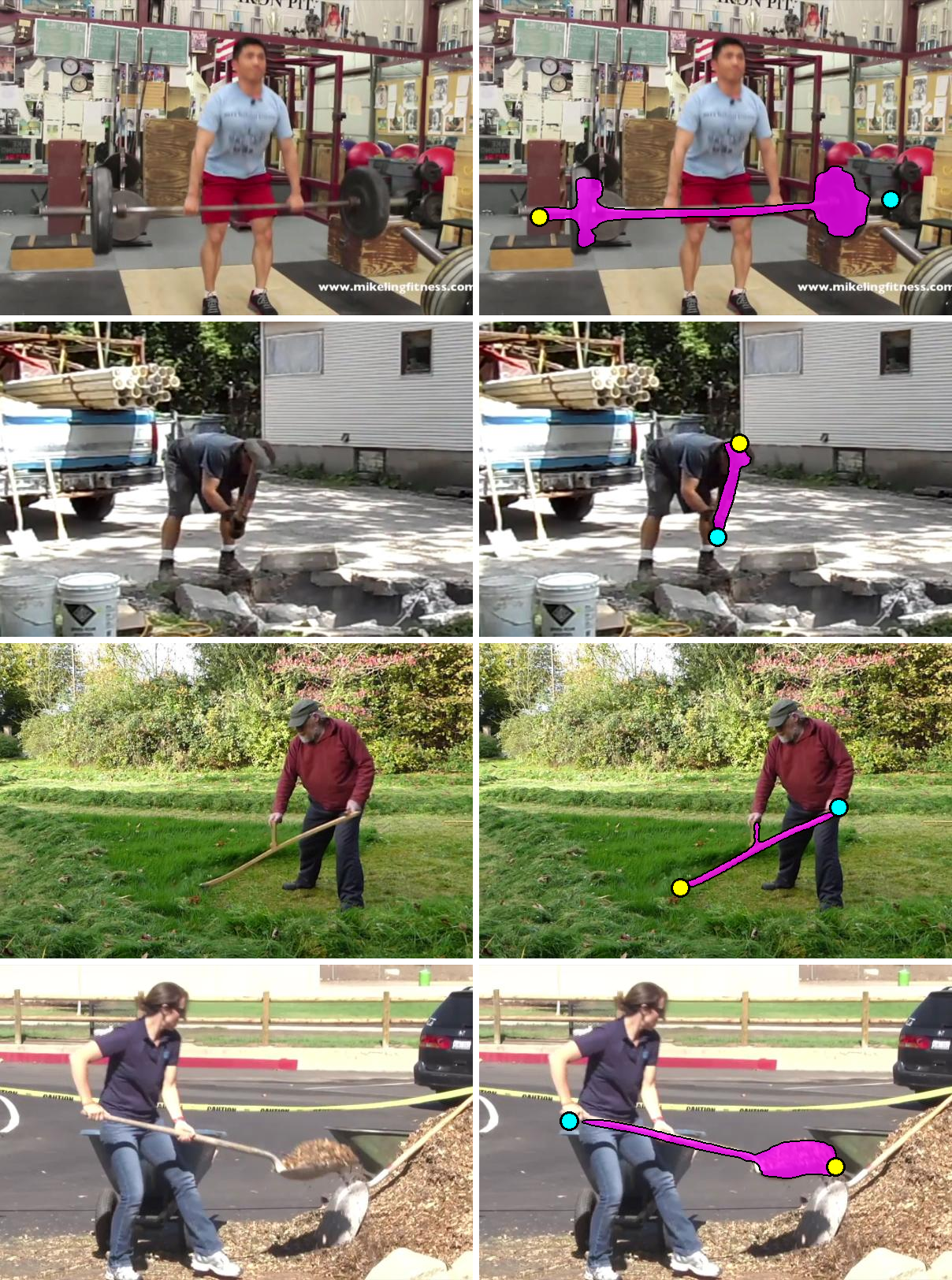}
    \caption{{\bf Detecting and localizing objects in video frames.} Example qualitative results on the Handtool dataset. \textbf{Left}: Input video frame (top to bottom: barbell, hammer, scythe, spade).
    \textbf{Right}: Output object mask (magenta) and object endpoints (yellow and cyan circles, corresponding to the ``head'' and the ``handle'' of the tool, respectively, where applicable).}
    \label{fig:endpoints_qualitative_results}
\end{figure}

At test time, we use the segmentation masks and bounding boxes from the trained Mask R-CNN to estimate the 2D coordinates of the object endpoints.
In our set-up, the Mask R-CNN is constrained to output no more than one segmented instance per image frame.
The endpoints are calculated as the intersection of a line fitted through the segmentation mask (estimate of object's main axis) and the bounding box (estimate of object's extremities).
However, we discard the endpoints if the distance of either wrist joint from the line segment between the endpoints is larger than a threshold (incorrect segmentation of the manipulated object).
The relative orientation of the object (i.e.\ which endpoint corresponds to the ``head'' of the tool and which to its ``handle'', for example)
is determined by the relative proximity of each endpoint to the wrist joints (hammer) or by the relative spatial location of the endpoints in the video frames (barbell, scythe, spade).
Figure \ref{fig:endpoints_qualitative_results} illustrates the output of our object localization and endpoint detection.

\begin{table*}[ht]
\centering
\begin{tabular}{lccccc}
\hline\noalign{\smallskip}
Method                     &   Kong-vault &  Muscle-up &  Pull-up  &     Safety-vault &    Avg \\ \noalign{\smallskip}\hline\noalign{\smallskip}
SMPLify~\citep{bogo2016keep}& 121.75 &   147.41 &   120.48  &  169.36 &  139.69 \\
HMR~\citep{hmrKanazawa18}&    111.36 &   140.16 &   132.44  &  149.64 &  135.65 \\
\citet{li2019motionforcesfromvideo}               & 98.42 & 125.21 & 119.92 & \underline{138.45}    & 122.11 \\
Ours (generic model)    & \underline{93.05} & \underline{124.55} & \underline{101.13}& 140.20    & \underline{116.13}  \\
\noalign{\smallskip}\hline\noalign{\smallskip}
Ours (action-specific models) & \textbf{92.77} & \textbf{122.83} & \textbf{99.98} & \textbf{137.32} &  \textbf{115.45}
\\  \noalign{\smallskip}\hline
\end{tabular}
\caption{Mean per joint position error (in mm) of the recovered 3D motion for each action on the Parkour dataset.}
\label{tb:mpjpe-galo-per-action}
\end{table*}

\section{Experiments}\label{sec:results}
In this section we present quantitative and qualitative evaluation of the reconstructed 3D person-object interactions.
Since we recover not only human poses but also object poses and contact forces, evaluating our results is difficult due to the lack of ground truth forces and 3D object poses in standard 3D pose benchmarks such as~\citet{h36m_pami}. 
Consequently, we evaluate our motion and force estimation quantitatively on a recent Biomechanics video/MoCap dataset capturing challenging dynamic parkour motions~\citep{maldonado}.
In addition, we report joint errors on our newly collected dataset of videos depicting handtool manipulation actions.
Furthermore, we show qualitative results on both datasets to demonstrate the quality of our motion/force estimation.
Finally, we discuss the main failure modes of our method at the end of the section.

\subsection{Parkour dataset}\label{sec:parkour_dataset}
This dataset contains RGB videos capturing  human subjects performing four typical parkour actions: kong-vault, moving-up, pull-up and safety-vault.
These are highly dynamic motions with rich contact interactions with the environment.
Half of the videos in the dataset are provided with ground truth 3D motion and contact forces captured with a Vicon motion capture system and force sensors.
Due to the blur of fast motion in the parkour actions, this dataset is challenging for computer vision algorithms.

\paragraph{Evaluation set-up.}
We evaluate our method on the 28 parkour sequences with ground truth 3D motion and contact forces, while the remaining videos are used for training the contact recognizer.
We evaluate the accuracy of the recovered 3D human poses using the common approach of computing the mean per joint position error~(MPJPE) of the estimated 3D pose with respect to the ground truth after rigid alignment~\citep{gower1975generalized}.
For evaluating contact forces we express the estimated and the ground truth 6D forces at the position of the contact aligned with the world coordinate frame provided in the dataset. We split the 6D force into linear and moment components and report the average Euclidean distance of the linear force and the moment with respect to the ground truth.

\begin{table*}[t]
\centering
\begin{tabular}{lcccccccc}
\hline
\noalign{\smallskip}\multirow{3}{*}{Method} & \multicolumn{2}{c}{L. Sole} & \multicolumn{2}{c}{R. Sole} & \multicolumn{2}{c}{L. Hand} & \multicolumn{2}{c}{R. Hand} \\  
                        & lin. force     & moment     & lin. force     & moment     & lin. force     & moment     & lin. force     & moment  \\ 
                        & (N)     & (N$\cdot$m)     & (N)     & (N$\cdot$m)     & (N)     & (N$\cdot$m)     & (N)     & (N$\cdot$m)  \\ \noalign{\smallskip}\hline\noalign{\smallskip}
\citet{li2019motionforcesfromvideo}     & 144.23         & 23.71      & 138.21         & 22.32      & 107.91         & 131.13     & 113.42         & 134.21     \\ 
Ours (generic model)           & \textbf{142.11}         & \textbf{22.91}      & \textbf{137.34}         & \textbf{20.11}      & \textbf{105.07}         & \textbf{130.42}     & \textbf{112.21}         & \textbf{132.94}     \\ \noalign{\smallskip}\hline
\end{tabular}
\caption{Estimation errors of the contact forces exerted on soles and hands on the Parkour dataset.}
\label{tb:force-errors-galo}
\end{table*}

\begin{table*}[t]
\centering
\begin{tabular}{lcccccc}
\hline\noalign{\smallskip}
Method                       & Barbell        & Spade         & Hammer         & Scythe         & Avg            \\ \noalign{\smallskip}\hline\noalign{\smallskip}
SMPLify~\citep{bogo2016keep}  & 130.69         & 135.03        & 93.43          & \textbf{112.93}         & 118.02         \\
HMR~\citep{hmrKanazawa18}     & 105.04         & 97.18         & 96.34          & 115.42         & 103.49         \\
\citet{li2019motionforcesfromvideo}               & 104.23 &  95.21&        95.87 & 114.22 & 102.38 \\
Ours (generic model) & \underline{83.95} & \underline{89.21}&       \underline{91.78} & 125.12 & \underline{97.51} \\
\noalign{\smallskip}\hline\noalign{\smallskip}
Ours (action-specific models) & \textbf{83.12} & \textbf{88.89} & \textbf{90.23} & \underline{114.13} & \textbf{94.09}
\\ \noalign{\smallskip}\hline
\end{tabular}
\caption{Mean per joint position error (in mm) of the recovered 3D human poses for each tool type on the Handtool dataset.}
\label{tb:mpjpe-handtools}
\end{table*}

\begin{table*}[t]
\centering
\begin{tabular}{lccccc}
\hline\noalign{\smallskip}
Method                     &   Kong-vault &  Muscle-up &  Pull-up  &     Safety-vault & Avg\\ \noalign{\smallskip}\hline\noalign{\smallskip}
Ours (without 3D data term) & 94.69 &   \textbf{124.12} &   103.87  &  141.88 & 117.55 \\
Ours (generic model)  & \textbf{93.05} & 124.55 & \textbf{101.13}& \textbf{140.20} & \textbf{116.13}
\\  \noalign{\smallskip}\hline
\end{tabular}
\caption{
Ablation of the 3D data term~\eqref{eq:data-3d}. 
We report the mean per joint position error (MPJPE) in mm of the estimated 3D human motion for each action on the Parkour dataset.
The first row corresponds to the ablated model, where the person 3D data term has been removed from the generic model. The second row corresponds to the generic model.
}
\label{tb:ablation-3d-data}
\end{table*}

\begin{table*}[t]
\centering
\begin{tabular}{lcccccccc}
\hline
\noalign{\smallskip}\multirow{3}{*}{Method} & \multicolumn{2}{c}{L. Sole} & \multicolumn{2}{c}{R. Sole} & \multicolumn{2}{c}{L. Hand} & \multicolumn{2}{c}{R. Hand} \\  
                        & lin. force     & moment     & lin. force     & moment     & lin. force     & moment     & lin. force     & moment  \\ 
                        & (N)     & (N$\cdot$m)     & (N)     & (N$\cdot$m)     & (N)     & (N$\cdot$m)     & (N)     & (N$\cdot$m)  \\ \noalign{\smallskip}\hline\noalign{\smallskip}
Ours (no force regularization)     & 148.74         & 86.54      & 144.12         & 79.45      & 128.55         & 137.45     & 133.79         & 136.43     \\ 
Ours (no $\|f_k\|^2$ in~\eqref{eq:contact_smooth})     & 143.76         & 23.78      & 139.60         & 22.19      & 109.10         & 133.29     & 117.89         & 133.87     \\ 
Ours (generic model)           & \textbf{142.11}         & \textbf{22.91}      & \textbf{137.34}         & \textbf{20.11}      & \textbf{105.07}         & \textbf{130.42}     & \textbf{112.21}         & \textbf{132.94}     \\ \noalign{\smallskip}\hline
\end{tabular}
\caption{
Ablation of force regularization terms (eq.~\eqref{eq:contact_smooth}). We report
estimation errors of contact forces exerted on soles and hands in the Parkour dataset.
The first row corresponds to the ablated model where both terms regularizing the temporal variation of the force and the force magnitude are removed from our generic model. 
The second row corresponds to the ablated model where the term regularizing the magnitude of the estimated force is removed.
The third row corresponds to our generic model.}
\label{tb:ablation-force}
\end{table*}

\begin{table*}[t]
\centering
\begin{tabular}{lcccc}
\hline\noalign{\smallskip}
Method                     & Barbell        & Hammer          & Scythe         & Spade       \\ 
\noalign{\smallskip}\hline\noalign{\smallskip}
Mask R-CNN \citep{MaskRCNN} & 33/42/54     & 35/44/45    & \textbf{63}/72/76       & 54/79/93    \\
Ours (generic model)        & \textbf{47/72/96}   & \textbf{63/91/98}   & 51\textbf{/87/98}    & \textbf{56/85/99} \\ \noalign{\smallskip}\hline
\end{tabular}
\caption{The percentage of endpoints for which the estimated 2D location lies within 25/50/100 pixels (in 600$\times$400 pixel image) from the manually annotated ground truth location. }
\label{table:evaluation2dposeall2550100}
\end{table*}

\begin{figure*}[t]
    \centering
    \includegraphics[width=\textwidth]{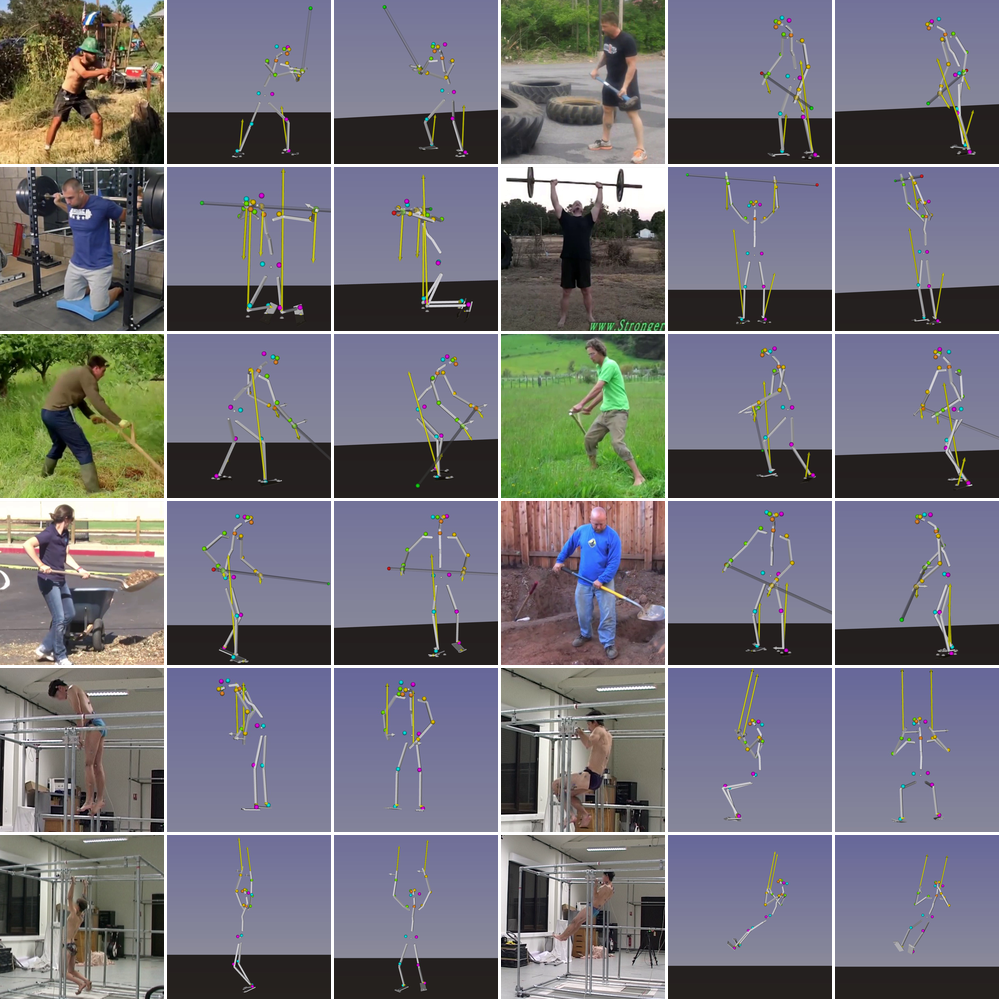}
    \caption{Example qualitative results on the Handtool (rows 1-4) and Parkour (rows 5-6) datasets.
    \textbf{Top-to-bottom}: hammer, barbell, scythe, spade, muscle-up and pull-up.
    Each example shows the input frame (left) and two different views of the output 3D pose of the person and the object (middle, right). 
    The yellow and the white arrows in the output show the contact forces and moments, respectively. 
    The length of the arrow represents the magnitude of the force normalized by gravity.
    }
    \label{fig:qualitative}
\end{figure*}

\begin{figure*}[t]
    \centering
    \includegraphics[width=\textwidth]{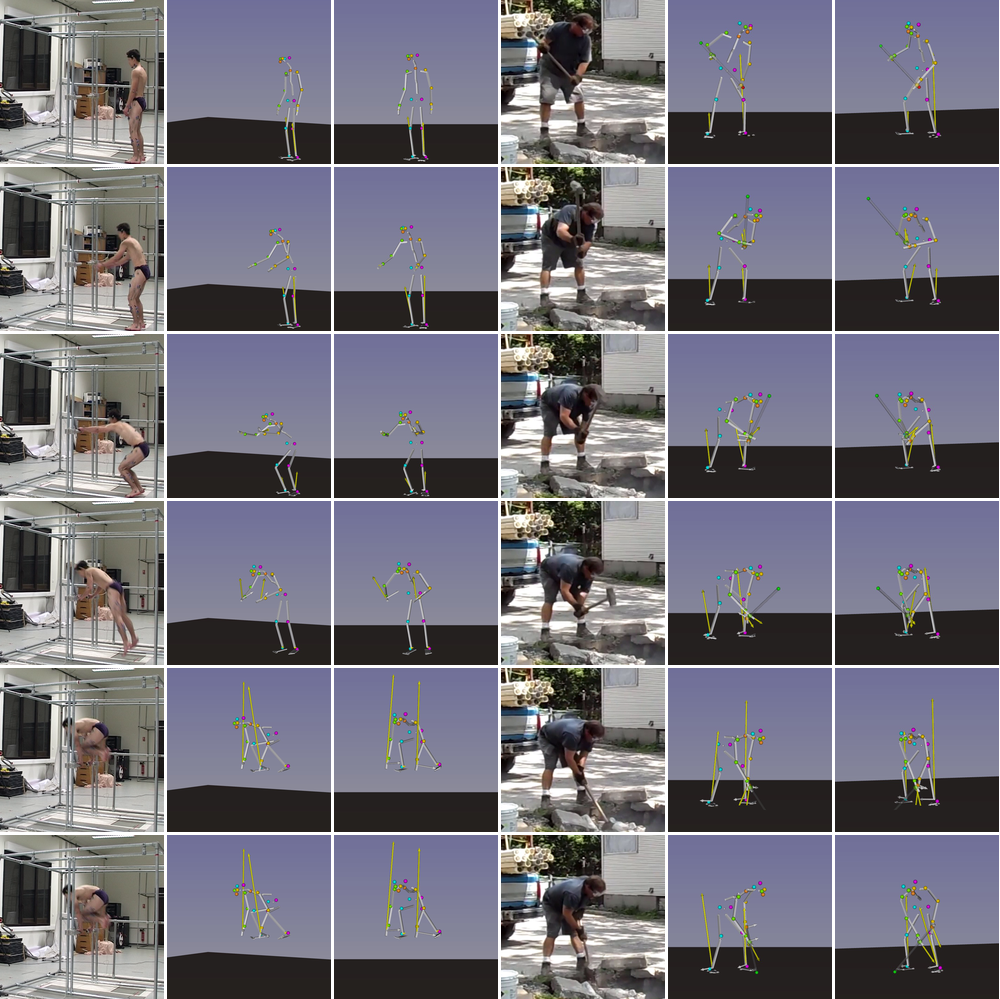}
    \caption{Example qualitative results on image sequences.
    \textbf{Columns 1-3}: muscle-up (Parkour dataset), \textbf{Columns 4-6}: hammer (Handtool dataset).
    {\bf Please see~\cite{project-page} for additional video results.}}
    \label{fig:sequence-results}
\end{figure*}

\paragraph{Results.}
We report joint errors for different actions in Table \ref{tb:mpjpe-galo-per-action} and compare results with the HMR~\citep{hmrKanazawa18} method, which is used to warm-start our method. 
To make it a fair comparison, we use the same Openpose 2D joints as input.
In addition, we evaluate the SMPLify \citep{bogo2016keep} 3D pose estimation method.
We also compare results with the previous version of this work~\citep{li2019motionforcesfromvideo}, which uses slightly different regularization of the estimated trajectory and forces.
We report results for two variants of our approach. The first variant (``generic model'') uses the same hyperparameters of the cost-function for all actions.
The second variant (``action-specific models'') uses action-specific hyperparameters adapted for each action (e.g. to regularize more strongly the motion of the legs in actions where legs are not used).
Starting from the hyper-parameters of the generic model, the action-specific hyper-parameters are obtained by performing grid search, as described in Section~\ref{sec:optimization} but here using validation videos of only one action class. 
The results show that our generic model outperforms all the baseline methods by more than 10mm on average on this challenging data, and that our action-specific models always achieve better performance on the corresponding actions compared to the baselines.

\begin{figure}[ht]
    \centering
    \includegraphics[width=0.49\textwidth]{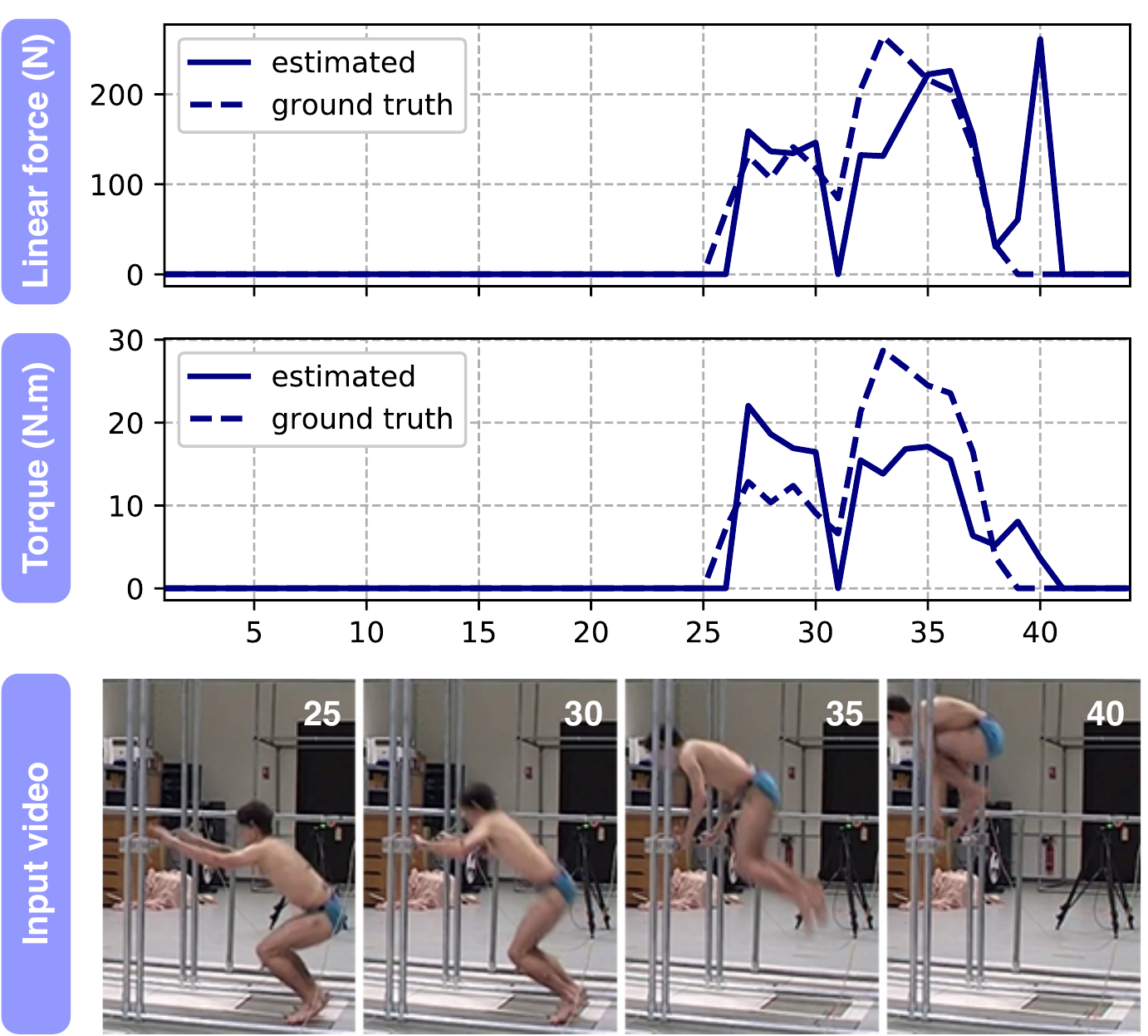}
    \caption{Example of temporal evolution of the magnitude of the estimated linear force (top) and torque (middle) at the person's left hand compared with the ground truth coming from the force sensors on an example sequence from the Parkour dataset. 
    The x-axis represents time (here frame numbers).
    Sample frames from the sequence with their corresponding frame numbers are shown at the bottom.}
    \label{fig:est-force-vs-gt}
\end{figure}

The force estimation results are summarized in Table \ref{tb:force-errors-galo} where we also report results of the previous version of this work~\citep{li2019motionforcesfromvideo}, which produces similar results.
We observe higher errors of the estimated moments at hands (compared to soles), which we believe is due to the challenging nature of the Parkour sequences where the entire person's body is often supported by hands.
In this case, the hand may exert significant force and torque to support the body, and a minor shift in the force direction may lead to significant errors.
In figure~\ref{fig:est-force-vs-gt}, we also show an example of temporal evolution of the magnitude of the estimated linear force and torque compared with the ground truth coming from the force sensors. 
The estimates correspond fairly well to the ground truth. We believe the spurious peak in the estimate around frame 40 is due to the error in contact recognition, which produces a spurious linear force and a small torque.

\subsection{Handtool dataset}\label{sec:handtooldataset}
In addition to the Parkour data captured in a controlled set-up, we would like to demonstrate generalization of our approach to the ``in the wild'' Internet instructional videos.
For this purpose, we have collected a dataset of object manipulation videos, which we refer to as the Handtool dataset.
The dataset contains videos of people manipulating four types of tools: {\em barbell}, {\em hammer}, {\em scythe}, and {\em spade}.
For each type of tool, we chose among the top videos returned by YouTube five videos covering a range of actions.
We then cropped short clips from each video showing the whole human body and the tool.

\paragraph{Evaluation of 3D human poses.}
For each video in the Handtool dataset, we have manually annotated the 3D positions of the person's left and right shoulders, elbows, wrist, hips, knees, and ankles, for the first, the middle, and the last frame.
The 3D annotation is done using the Berkeley Human Annotation Tool~\citep{humanannotationtool}, by following these three steps:
(i) annotate the 2D joint locations in the image, (ii) specify the relative depth ordering for linked joints, and (iii) run the optimization approach described in~\citet{taylor2000reconstruction} to obtain a 3D stick figure. This annotation process is repeated until the 3D figure is visually correct according to the annotator.
We evaluate the accuracy of the recovered 3D human poses by computing their MPJPE after rigid alignment.
Quantitative evaluation of the recovered 3D poses is shown table \ref{tb:mpjpe-handtools}.
On average, our generic model (the same as for the Parkour dataset) outperforms all the baselines on this dataset.
Our action-specific models achieve on average even better performance.
Our approach achieves the best results on all individual actions except on scythe. After manual inspection of the results,  we believe that this is due to the inaccuracy of the 3D model of the scythe, which is represented as a 3D line segment without explicitly modelling the handle of the scythe, which in turn affects the accuracy of the estimated 3D human poses (via the person-object contact model).
However, the differences between the methods are reaching the limits of the
accuracy of the manually provided 3D human pose annotations on this dataset.
For example,~\citet{marinoiu2013pictorial} point out that manual 3D annotation errors can range up to 100 mm per joint~\citep{h36m_pami}.

\paragraph{Evaluation of 2D object poses.}
To evaluate the quality of estimated object poses, we manually annotated 2D object endpoints in every 5th frame of each video in the Handtool dataset and calculated the 2D Euclidean distance (in pixels) between each manually annotated endpoint and its estimated 2D location provided by our method.
The 2D location is obtained by projecting the estimated 3D tool position back to the image plane. 
We compare our results to the output of the Mask R-CNN instance segmentation baseline~\citep{MaskRCNN} (which provides initialization for our person-object interaction model).
In Table~\ref{table:evaluation2dposeall2550100} we report for both methods the percentage of endpoints for which the estimated endpoint location lies within 25, 50, and 100 pixels from the annotated ground truth endpoint location. 
The results demonstrate that our approach provides in most cases more accurate and stable object endpoint locations compared to the Mask R-CNN baseline thanks to modeling the interaction between the object and the person. Lower results of our approach for scythe for the strict 25 pixel threshold can be again attributed to the inaccuracy of the 3D scythe model approximated only as a 3D line segment. 

\subsection{Ablation study}\label{sec:ablation-study}
To gain further insight into the improvements over the conference version of this work~\citep{li2019motionforcesfromvideo}, we perform an ablation study of (i) the newly introduced person 3D consistency loss~\eqref{eq:data-3d} (also referred to as the 3D data term) and (ii) 
the new force regularization term~\eqref{eq:contact_smooth}, which smooths not only the temporal variation but also the magnitude of the estimated contact forces.
These experiments are done using the Parkour dataset which has precise and dense ground truth for the 3D motion and contact forces captured by MoCap and force sensors.
Unless otherwise mentioned, the experiments are based on the generic model described previously.

\paragraph{Ablation of the 3D data term.}
In this ablation, we remove the 3D data term~\eqref{eq:data-3d} from the generic model while keeping the rest of the cost terms and the related parameters.
The results are reported in Table~\ref{tb:ablation-3d-data}, where we compare the mean per joint position error (MPJPE) of the ablated model with the original generic model. The results show that on average the new 3D data term improves the 3D pose estimates, though the improvement is relatively minor. Qualitatively, we have observed that the 3D data term plays the role of a pose prior that encodes, for example, the relative depth of the different joints (e.g. between the person's left and right hand), which is captured in the strong 3D prior of the HMR approach~\citep{hmrKanazawa18}.

\begin{figure*}[ht]
    \centering
    \includegraphics[width=0.94\textwidth]{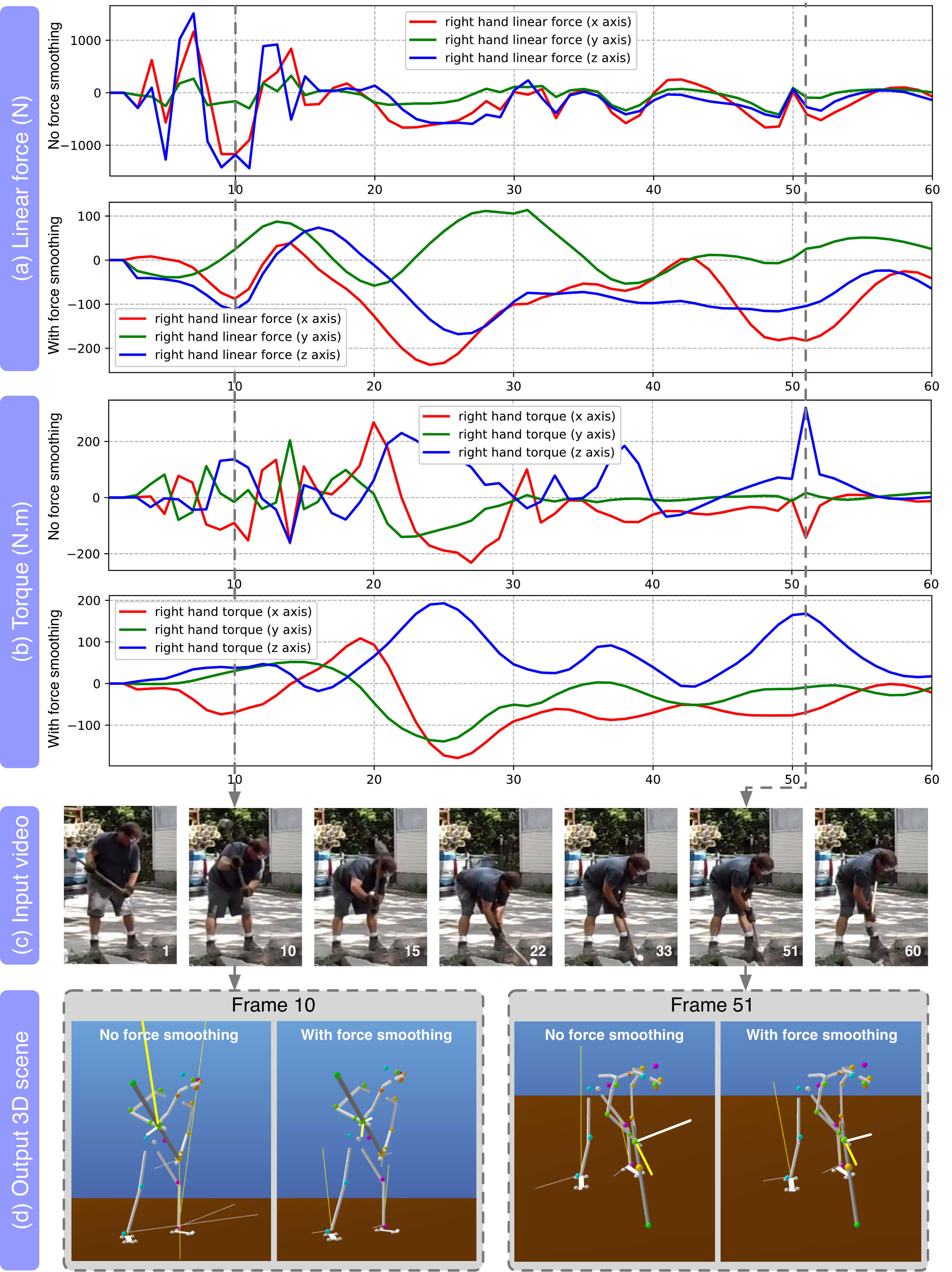}
    \caption{Plots of the estimated linear contact force {\bf(a)} and torque {\bf(b)} at the right hand for an example video from the Handtool dataset. In all plots the x-axis represents time (in frame numbers). In both (a) and (b), the top plot is without force regularization and the bottom plot is with force regularization (i.e. the generic model). {\bf(c)} shows example frames with their corresponding frame numbers. {\bf(d)} shows the estimated 3D scene at two sample frames with the highlighted linear contact force (bold yellow) and torque (bold white) at the right hand. Please note how force regularization effectively smoothes the estimated forces and torques reducing their unrealistic abrupt temporal changes and large magnitudes (note the different scales of the y-axis in the different plots).
    }
    \label{fig:force-plot}
\end{figure*}

\paragraph{Ablation of force regularization.}
The new force regularization term~\eqref{eq:contact_smooth} smooths not only the temporal variation of the estimated contact forces~\citep{li2019motionforcesfromvideo} but also the magnitude of the estimated contact forces.
Therefore, we evaluate and compare two ablated models against our generic model.
In the first ablated model (Ours (no force regularization)), we remove both terms regularizing the temporal variation and the magnitude of the estimated contact forces (i.e. the second and the third term in Eq.~\eqref{eq:contact_smooth}).
In the second ablated model (Ours (no $\|f_k\|^2$ in~\eqref{eq:contact_smooth})), we only remove the third term regularizing the magnitude of the estimated contact forces, i.e. this model regularizes only the temporal variation of the estimated contact forces. Note that this form of force regularization was used in the conference version of this work~\citep{li2019motionforcesfromvideo}.
Quantitative results are reported in Table~\ref{tb:ablation-force} and clearly show the benefit of regularizing both the temporal variation and the magnitude of the estimated contact forces (Ours (generic model)), which results in the lowest errors.
While we cannot compute force estimation errors on the Handtool dataset due to the lack of ground truth data, we can still perform a simple ablation analysis by plotting the temporal variation of the estimated linear forces and torques with and without the force regularization terms. This is shown on an example video sequence for the left-hand contact force in~Figure~\ref{fig:force-plot}.  
Please note how the regularization of both the temporal variation and magnitude of the estimated forces~\eqref{eq:contact_smooth} effectively smoothes the estimated forces reducing their abrupt temporal changes and unrealistic magnitudes.
Figure~\ref{fig:force-plot}(d) also compares the output of our model with and without force regularization at two example frames. In particular, frame \#10 corresponds to the case where the model without force regularization outputs a linear force with an unrealistic orientation and magnitude (highlighted with a bold yellow line in the image) whereas the regularized model outputs a more realistic force estimate in terms of both the orientation and magnitude. Similarly, for frame \#51  the model with force regularization outputs a torque with a smaller and hence more realistic magnitude.  

\begin{figure*}[t]
    \centering
    \includegraphics[width=\textwidth]{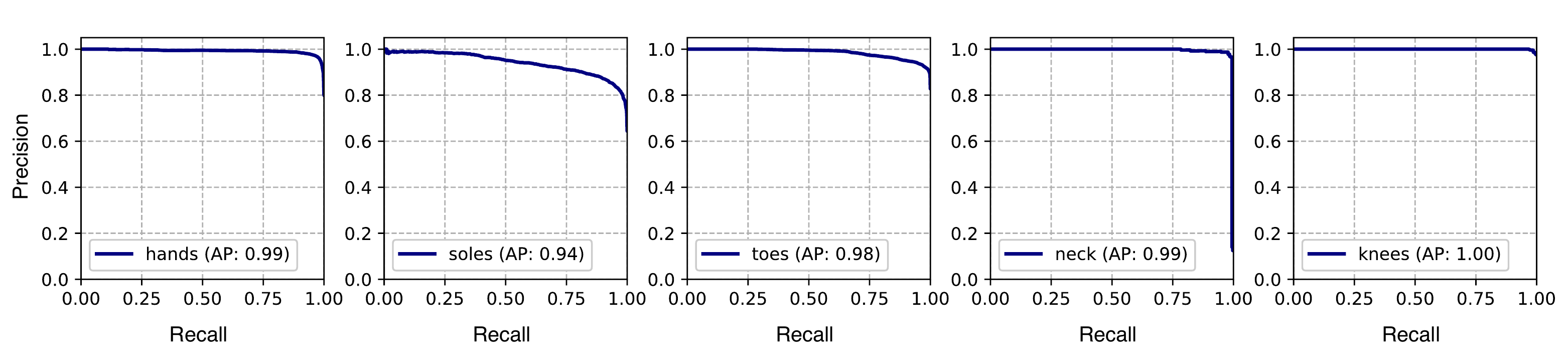}
    \caption{Precision-recall curves of our trained models for recognizing the hand-, sole-, toes-, neck-, and knee-contact state. }
    \label{fig:pr-curve}
\end{figure*}

\subsection{Evaluation of contact recognition}
In this section we evaluate the quality of our contact recognizers. 
The training data, the recognition architecture and the training process are described in Section~\ref{sec:contact_recognition}.
To form the test set we annotate contact states in the entire Handtool dataset and a subset of the Parkour dataset obtained by sampling every 5-th frame.
Following the same annotation process as done for training, we have cropped image patches around individual human joints in the test set. This results in a separate test set for each of the five joint types: hand, sole, toes, neck and knee. The neck and the knee test sets include only patches from the Handtool dataset as the Parkour dataset does not consider these types of contacts.
We evaluate each contact recognizer using a precision-recall curve on its corresponding test set. The positive class means the joint is ``in contact".  The evaluation results are shown in Fig.~\ref{fig:pr-curve}. Each precision-recall curve is also summarized using average precision (AP). The results demonstrate good quality of our contact recognition models despite the appearance variation present in both the Handtool and Parkour datasets.

\subsection{Qualitative results}\label{sec:qualitative-results}
Here we show qualitative examples.
Additional video results are available on our~\citet{project-page}.

Figure~\ref{fig:qualitative} shows a collection of qualitative results at sampled video frames in the Handtool (top four rows) and the Parkour (bottom four rows) datasets.
For each sample, we first show the original frame (left image), followed by the estimated 3D motion and forces from the original viewpoint (middle image), and the same 3D scene from a different viewpoint (right image).
Note that for the Parkour dataset we recognize the contact states of human joints but do not recognize and model the pose of the object (the metal construction) the person is interacting with.
In addition to results for individual frames from different videos, we provide in Figure~\ref{fig:sequence-results} results for two sequences of frames to demonstrate the continuity of the reconstructed actions.
The sequences demonstrate that the outputs of our method are temporally consistent and smooth.

\begin{figure*}[ht]
    \centering
    \includegraphics[width=0.99\textwidth]{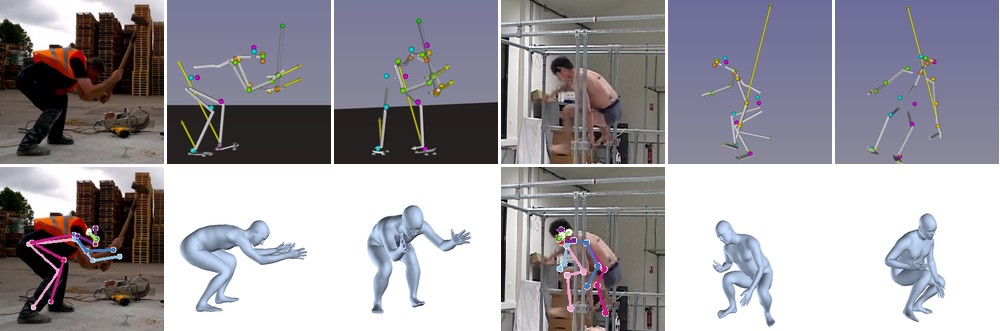}
    \caption{Qualitative comparison with the baseline HMR estimator~\citep{hmrKanazawa18}. In each example, the top row shows the input frame (left) and the output of our method from two different viewpoints (middle, right). The bottom row shows the estimated 2D joints (left) and the output of the HMR baseline shown from two different viewpoints (middle, right).
    In the hammering example (the left panel) the person's hands holding the hammer are restricted to be on the handle by our contact model, thus reducing the depth ambiguity compared to 3D human poses provided by the baseline HMR~\citep{hmrKanazawa18} estimator,  which often outputs open arms.
    The second example (the right panel) shows an ``outlier'' frame of a Parkour video where HMR fails to estimate correct human body orientation w.r.t the camera due to heavy occlusion and motion blur.
    }
    \label{fig:comparison-hmr}
\end{figure*}

\begin{figure}[ht]
    \centering
    \includegraphics[width=0.49\textwidth]{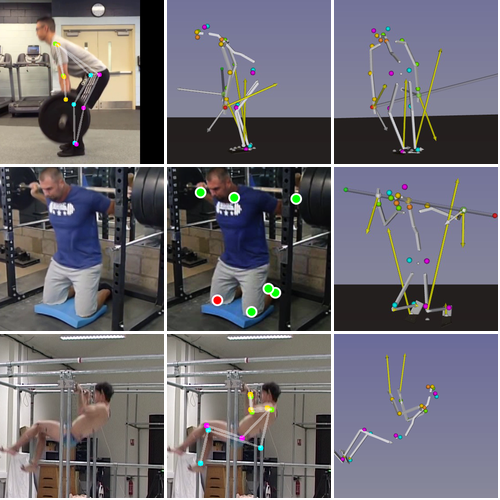}
    \caption{Main failure modes of our method: (i) missing object 2D endpoint detections (top row): the handle of the barbell is not detected, which affects the 3D output of our model; (ii) contact recognition errors (middle row): the person's right knee is incorrectly recognized as not in contact (red), leading to incorrect force estimates shown on the right;
    (iii) incorrect 2D human joints (bottom row): the missing 2D detection of the person's left foot has lead to errors in estimating the 3D location of the left leg. }
    \label{fig:failure_modes}
\end{figure}

Figure~\ref{fig:comparison-hmr} shows a comparison of our model with the baseline HMR approach~\citep{hmrKanazawa18}. 
In the first example (hammering action), the person's hands holding the hammer are restricted to be on the handle by our contact model, thus reducing the depth ambiguity compared to 3D human poses provided by the baseline HMR~\citep{hmrKanazawa18} estimator,  which often outputs open arms.
The second example shows an ``outlier'' frame of a Parkour video where HMR fails to estimate correct human body orientation due to heavy occlusion and motion blur.
In this case, our method relies on the model of dynamics and the pose prior to synthesize the person's motion in between good predictions.
Due to these reasons, we observe that our method often predicts better poses than the baseline methods that are applied to individual frames and do not model the temporal interaction between the person and the tool.

The qualitative results also demonstrate that our model predicts reasonable contact forces.
The directions of the contact forces exerted on the person's hands are consistent with the object's motion trajectory and gravity, and the ground reaction forces generally point towards the direction opposite to gravity.
Specifically, in the video with the person practicing back squat with barbell (see the left example in the second row of Figure~\ref{fig:qualitative}), the reconstructed object contact forces and ground reaction forces are distributed evenly on the person's hands, and knees, respectively.
Another example is scythe  (third row of Figure~\ref{fig:qualitative}, right), where the distribution of ground reaction forces at the person's feet follows the swings of the body while cutting the grass.
In the shown frame the person's center of mass is above their right leg, leading to larger contact force at the right leg.

\subsection{Failure modes}
 Figure~\ref{fig:failure_modes} shows three typical failure modes described below.

\paragraph{Missing object 2D endpoint detections.} 
The estimated object 2D endpoints are often noisy due to heavy occlusions between human limbs and the manipulated object.
To solve this problem, we filter out  endpoints with low confidence at the end of the recognition stage.
However, this produces missing observations in the estimated 2D endpoint sequences as shown in the first example in Figure~\ref{fig:failure_modes}, where there is no predicted endpoint as the barbell handle is completely occluded (imaged from the side).
In this case, our contact motion model can infer the position of the barbell handle from the position of hands, but the results are often not very accurate.

\paragraph{Contact recognition errors.} 
The second row of Figure~\ref{fig:failure_modes} shows an example with incorrectly estimated contact state. In this case, the person's right knee is incorrectly recognized as not in contact, leading to incorrect force estimation.

\paragraph{Incorrectly localized human joints in the image.}
Our method struggles to estimate correct 3D poses if the quality of 2D detection is low.
An example is shown in the bottom row of Figure~\ref{fig:failure_modes}, where the missing 2D detection of the person's left foot has lead to errors in estimating the 3D location of the left leg.

\subsection{Limitations}
Our approach has several limitations, which we discuss next. 
First, our object model is currently limited to rigid, stick-like tools. Modeling other types of rigid objects, e.g. boxes, would require recognizing and modelling other object shapes, which is technically possible with our model but we leave it for future work. Recognizing and modelling interactions with non-rigid objects such as cloth is still an open challenge.   
Second, the proposed method models the hand-object contact at a relatively coarse level by taking into account only a single joint location (the wrist). While this is reasonable for the type of objects considered in this work, it is too coarse for a more fine-grained manipulation of smaller objects such as pencils or cups.
Third, we initialize our model with~\citep{hmrKanazawa18} to provide the size and shape of the depicted person, but then use only the body skeletal rig in the estimation stage. A mesh-based representation could be more descriptive.  
Finally, our method does not consider object-object and object-ground interactions. For example, in the case of breaking concrete with a hammer, our method does not currently model the contact force exerted on the hammer by the concrete. Modeling the interactions between the object and the environment is an exciting direction of future work.

\section{Conclusion}
We have developed a visual recognition system that takes as input video frames together with a simple object model, and outputs a 3D motion of the person and the object including contact forces and torques actuated by the human limbs. 

We have validated our approach on a recent video MoCap dataset with ground truth contact forces.
Finally, we have collected a new dataset of unconstrained instructional videos depicting people manipulating different objects and have demonstrated benefits of our approach on this data. 
Our work opens up the possibility of large-scale learning of human-object interactions from Internet instructional videos~\citep{Alayrac16unsupervised}.

\begin{acknowledgements}
We thank Bruno Watier (Universit\'e Paul Sabatier and LAAS-CNRS) and Galo Maldonado (ENSAM ParisTech) for making public the Parkour dataset.
This work was partly supported by the ERC grant LEAP (No. 336845), the French government under management of Agence Nationale de la Recherche as part of the “Investissements d’avenir” program, references ANR-19-P3IA-0001 (PRAIRIE 3IA Institute) and ANR-19-P3IA-0004 (ANITI 3IA Institute) , and the European Regional Development Fund under the project IMPACT (reg. no. CZ.02.1.01/0.0/0.0/15 003/0000468).
\end{acknowledgements}

\clearpage

\bibliographystyle{spbasic}      
\bibliography{references}   

\clearpage
\appendix

\section*{Outline of the appendix}
In this appendix, we provide additional technical details of the proposed method.
In appendix~\ref{appendix:human_object_models}, we provide a comprehensive description of the parametric human and object model we use for the trajectory optimization.
Then, in appendix~\ref{appendix:ground_force_generators} we give details of the ground contact force generators mentioned in the main paper (section~\ref{sec:physical_plausibility}).

\section{Parametric human and object models}
\label{appendix:human_object_models}
\begin{figure}[t]
    \centering
    \includegraphics[width=0.35\textwidth]{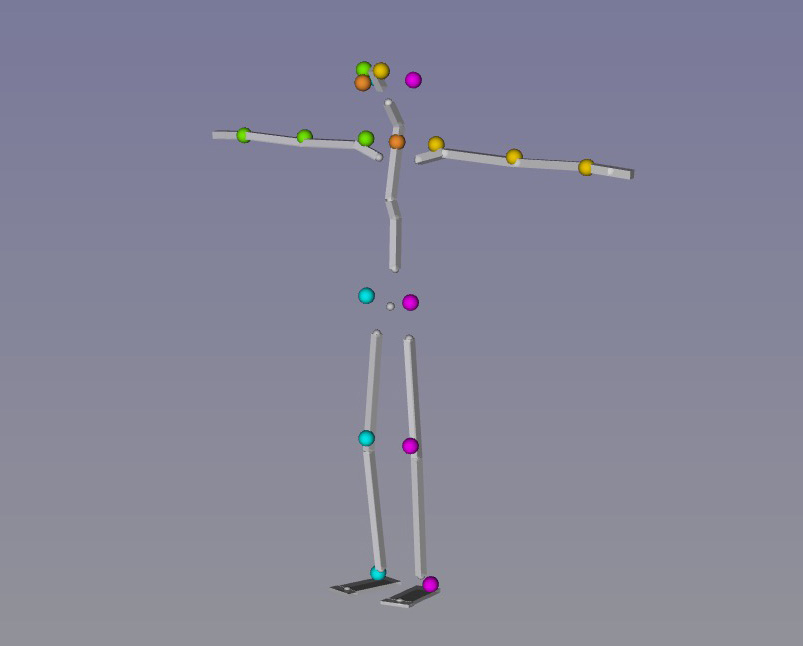}
    \caption{Our human model in the reference posture. The skeleton consists of one free-floating basis joint corresponding to pelvis, and 23 spherical joints. The colored spheres are 18 virtual markers that correspond to 18 OpenPose joints. Each marker is associated to a semantic joint in our model.}
    \label{fig:human_model}
\end{figure}

\paragraph{Human model.} 
We model the human body as a multi-body system consisting of a set of rotating joints and rigid links connecting them.
We adopt the joint definition of the SMPL model~\citep{loper2015smpl} and approximate the human skeleton as a kinematic tree with 24 joints: one free-floating joint and 23 spherical joints.
Figure \ref{fig:human_model} illustrates our human model in a canonical pose.
A free-floating joint consists of a 3-dof translation in $\mathbb{R}^3$ and a 3-dof rotation in $SO(3)$; we model the pelvis by a free-floating joint to describe the person's body orientation and translation in the world coordinate frame.
A spherical joint is a 3-dof rotation; it represents the relative rotation between two connected links in our model.
In practice, we use unit quaternions to represent 3D rotations and axis-angles to describe angular velocities.
As a result, the configuration vector of our human model $q^\mathrm{h}$ is a concatenation of the configuration vectors of the 23 spherical joints (dimension 4) and the free-floating pelvis joint (dimension 7), hence of dimension 99.
The corresponding human joint velocity $\dot{q}^\mathrm{h}$ is of dimension $23\times 3+6=75$ (by replacing the quaternions with axis-angles).
For simplicity, in the main paper we do not distinguish this difference in dimension and consider both $q^\mathrm{h}$ and $\dot{q}^\mathrm{h}$ to be represented using axis-angles, hence of the same dimension $n_q^\mathrm{h}=75$.
In addition, based on these 24 joints, we define 18 ``virtual markers'' (shown as colored spheres in Figure \ref{fig:human_model}) that represent the 18 OpenPose joints.
These markers are used instead of the 24 joints to compute the re-projection errors with respect to the OpenPose 2D detections.

\begin{figure}[t]
    \centering
    \includegraphics[width=0.25\textwidth]{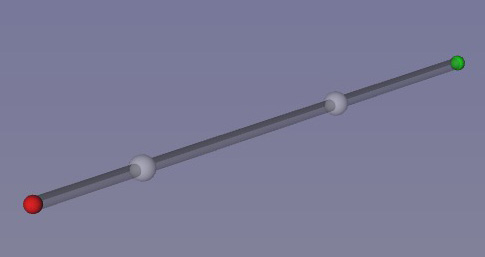}
    \caption{All four handtools are represented by a single object model shown in this image.
    The object model consists of 1 free-floating basis joint corresponding to the handle end point (red sphere), 1 prismatic joint corresponding to the head of the tool (green sphere), and several prismatic joints corresponding to the location of the contact points (grey translucent spheres in the middle).
    The contact points should lie on the feasible contact area (grey stick) formed by the two endpoints.}
    \label{fig:object_model}
\end{figure}

\paragraph{Object models.}
All four objects, namely barbell, hammer, scythe and spade, are modeled as a non-deformable rigid line stick.
The configuration $q^\mathrm{o}$ represents the 6-dof displacement of the stick handle, as illustrated in Figure \ref{fig:object_model}.
In practice, $q^\mathrm{o}$ is a 7-dimensional vector containing the 3D translation and 4D quaternion rotation of the free-floating handle end.
The object joint velocity $\dot{q}^\mathrm{o}$ is of dimension 6 (by replacing the quaternion with an axis-angle).
The handtools that we are modelling have the stick handle as the contact area. 
We ignore the handle's thickness and represent the contact area using the line segment between the two endpoints of the handle.
Depending on the number of human joints in contact with the object, we associate the same number of contact points to the object's local coordinate frame.
These contact points can be located at any point along the feasible contact area.
In practice, all object contact points together with the endpoint corresponding to the head of the handtool are implemented as ``virtual'' prismatic joints of dimension 1.


\section{Generators of the ground contact forces}
\label{appendix:ground_force_generators}

In this section, we describe the generators $g^{(3)}_n$ and $g^{(6)}_{kn}$ for computing the contact forces exerted by the ground on the person.
Recall from the main paper that we consider different contact models depending on the type of the joint. We model the planar contacts between the human sole and the ground by fitting the point contact model (given by Eq.~\eqref{eq:point_contact} in the main paper) at each of the four sole vertices.
For other types of ground contacts, e.g.~the knee-ground contact, we apply the point contact model directly at the human joint.
We model the ground as a 2D plane $G = \{p\in \mathbb{R}^3|a^Tp=b\}$ with a normal vector $a\in \mathbb{R}^3$, $a\neq 0$, $b\in \mathbb{R}$ and a friction coefficient $\mu$.
In the following, we first provide the expression of the 3D generators $g^{(3)}_n$ for modeling point contact forces and then derive the 6D generators $g^{(6)}_{kn}$ for modeling planar contact forces.

\paragraph{3D generators $g^{(3)}_n$ for point contact forces.}
Let $p_k$ be the position of a contact point $k$ located on the ground surface, i.e. $a^Tp_k=b$.
We define at contact point $k$ a right-hand coordinate frame $C$ whose $xz$-plane overlaps the plane $G$ and whose $y$-axis points towards the gravity direction, i.e., the opposite direction to the ground normal $a$.
During point contact, it is a common assumption that the ground exerts only linear reaction forces on the contact point $c$.
In other words, the spatial contact force expressed in the local frame $C$ can be expressed as
\begin{align}
    ^C\phi = 
    \begin{pmatrix}
        f \\ 
        \mathbf{0}_{3\times1}
    \end{pmatrix}, \label{eq:point_contact_force}
\end{align}
where the linear component $f$ must lie in the second-order cone $\mathcal{K}^3 = \{f=(f_x,f_y,f_z)^T|\sqrt{f_x^2 + f_z^2} \leq -f_y \tan\mu\}$, which can be approximated by the pyramid ${\mathcal{K}^3}^\prime = \{f=\sum_{n=1}^4{\lambda_n g^{(3)}_n}|\lambda_n\geq 0\}$, with a set of 3D-generators
\begin{align}
    g^{(3)}_1 &= 
    \left(\sin{\mu}, -\cos{\mu}, 0\right)^T, \\
    g^{(3)}_2 &= 
    \left(-\sin{\mu}, -\cos{\mu}, 0\right)^T, \\
    g^{(3)}_3 &= 
    \left(0, -\cos{\mu}, \sin{\mu}\right)^T, \\
    g^{(3)}_4 &= 
    \left(0, -\cos{\mu}, -\sin{\mu}\right)^T,
\end{align}
where $\mu$ is the friction coefficient. 
More formally, we are approximating the friction cone $\mathcal{K}^3$ with the conic hull ${\mathcal{K}^3}^\prime$ spanned by 4 points on the boundary of $\mathcal{K}^3$, namely, $g^{(3)}_n$ with $n=1,2,3,4$. 

\paragraph{6D generators $g^{(6)}_{kn}$ for planar (sole) contact forces.}
Here we show how to obtain the 6D generator $g^{(6)}_{kn}$ from $g^{(3)}_{n}$ and the contact point position $p_k$.
As described in the main paper, we approximate human sole as a rectangle area with 4 contact points.
We assume that the sole overlaps the ground plane $G$ during contact.
Similar to the point contact, we define 5 parallel coordinate frames $C_k$, one at each of the four sole contact points, plus a frame $A$ at the ankle joint.
Note that the frames $C_k$ and $A$ are parallel to each other, i.e., there is no rotation but only translation when passing from one frame to another.
We can write the contact force at contact point $k$ as the 6D spatial force
\begin{align}
    ^{C_k}\phi_k =
    \sum_{n=1}^4 \lambda_{kn}
    \begin{pmatrix}
        g^{(3)}_n \\ 
        \mathbf{0}_{3\times1}
    \end{pmatrix}
    , \text{ with } \lambda_{kn} \geq 0.
\end{align}
We denote by $^Ap_k$ the position of contact point $c_k$ in the ankle frame $A$, and by $^AX_{C_k}^*$ the matrix converting spatial forces from frame $C_k$ to frame $A$.
We can then express the contact force in frame $A$:
\begin{align}
    ^A\phi &= \sum_{k=1}^4{{^AX_{C_k}^*}^{C_k}\phi_k} \\
    &= \sum_{k=1}^4{\begin{pmatrix}
        I_3 & ^Ap_k\times\\
        0_3 & I_3\\
    \end{pmatrix}^{-T}{^{C_k}\phi_k}} \\
    &=\sum_{k=1}^4\sum_{n=1}^4 \lambda_{kn} g^{(6)}_{kn}, \label{eq:planar_contact_force_6d}
\end{align}
where 
\begin{align}
    g^{(6)}_{kn}=
    \begin{pmatrix}
        g^{(3)}_n \\ 
        ^Ap_k\times g^{(3)}_n
    \end{pmatrix}.
\end{align}

\end{document}